\documentclass[10pt,twocolumn,letterpaper]{article}
\usepackage{cvpr}              

\usepackage[accsupp]{axessibility} 
\usepackage{graphicx}
\usepackage{amsmath}
\usepackage{amssymb}
\usepackage{booktabs}
\usepackage{times}
\usepackage{epsfig}
\usepackage{graphicx}
\usepackage{amsmath}
\usepackage{amssymb}
\usepackage{helvet}
\usepackage{courier}
\usepackage{upgreek}
\usepackage{color}
\usepackage{url}
\usepackage{cite}
\usepackage{multirow}
\usepackage{diagbox}
\usepackage{bm}
\usepackage{cases}
\usepackage{colortbl}
\usepackage{amsfonts}

\usepackage[dvipsnames]{xcolor}

\usepackage[pagebackref,breaklinks,colorlinks]{hyperref}

\usepackage[capitalize]{cleveref}
\crefname{section}{Sec.}{Secs.}
\Crefname{section}{Section}{Sections}
\Crefname{table}{Table}{Tables}
\crefname{table}{Tab.}{Tabs.}

\def\cvprPaperID{2027} 
\def\confName{CVPR}
\def\confYear{2022}

\begin{document}

\title{Rope3D: The
Roadside Perception Dataset for Autonomous Driving \\ and Monocular 3D Object Detection Task}


\author
{ 
Xiaoqing Ye\textsuperscript{1}\thanks{indicates equal contribution} 
\quad
Mao Shu\textsuperscript{1}$^*$
\quad
Hanyu Li \textsuperscript{1}
\quad
Yifeng Shi\textsuperscript{1}\\
\quad
Yingying Li\textsuperscript{1}
\quad
Guangjie Wang\textsuperscript{2}
\quad
Xiao Tan\textsuperscript{1}\thanks{Xiao Tan (tanxchong@gmail.com is the corresponding author.)}
\quad
Errui Ding \textsuperscript{1}
\\[1em]
 \\ 
 \textsuperscript{1}Baidu Inc.
\quad
 \textsuperscript{2}China University of Mining and Technology
}

\maketitle
\begin{abstract}
 Concurrent perception datasets for autonomous driving are mainly limited to frontal view with sensors mounted on the vehicle. None of them is designed for the overlooked roadside perception tasks. On the other hand, the data captured from roadside cameras have strengths over frontal-view data, which is believed to facilitate a safer and more intelligent autonomous driving system. To accelerate the progress of roadside perception, we present the first high-diversity challenging \textbf{Ro}adside \textbf{Pe}rception 3D dataset- \textbf{Rope3D} from a novel view. The dataset consists of 50k images and over 1.5M 3D objects in various scenes, which are captured under different settings including various cameras with ambiguous mounting positions, camera specifications, viewpoints, and different environmental conditions. We conduct strict 2D-3D joint annotation and comprehensive data analysis, as well as set up a new 3D roadside perception benchmark with metrics and evaluation devkit. Furthermore, we tailor the existing frontal-view monocular 3D object detection approaches and propose to leverage the geometry constraint to solve the inherent ambiguities caused by various sensors, viewpoints. Our dataset is available on \href{https://thudair.baai.ac.cn/rope}{
https://thudair.baai.ac.cn/rope}.
\end{abstract}

\begin{figure}[t]
 \begin{center}
  \includegraphics[width=\linewidth]{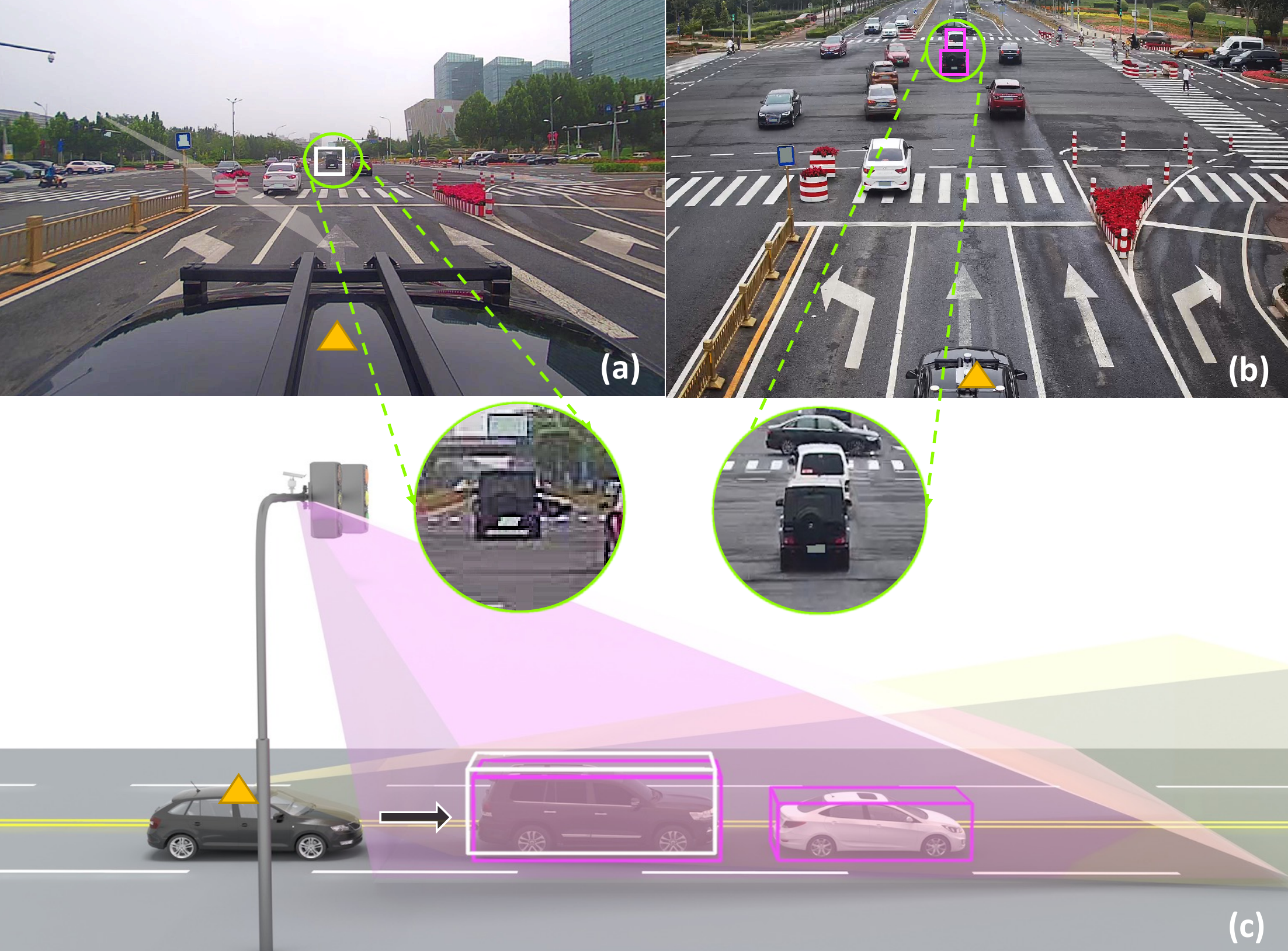}
 \end{center}
 \caption{The comparison of (a) frontal view and (b) roadside camera view with a pitch angle. The car view focuses more on the frontal area whereas the roadside camera observes the scene in a long-term and large-range manner. Vehicles can be easily occluded by closer objects in frontal view but the roadside view alleviates the risk. 
 For example,
 for car-view (a), the white van is occluded by the black jeep whereas in roadside view (b) they are both visible, corresponding to the white and pink 3D boxes in (c). The triangle mark denotes the same LiDAR-mounted vehicle.}
 \label{fig:different_view}
\end{figure}
\section{Introduction}
\label{sec:intro}
Autonomous driving plays a crucial role in helping reduce traffic accidents and improve transportation efficiency. Current perceptual systems mainly equip the moving vehicle with LiDAR or camera sensors. Owing to the movement, the vehicle perceptual system can not observe surroundings for a long period. In addition, since the mounted sensor is relatively low (usually on the top of a vehicle), the perceptual range is comparatively limited and is vulnerable to occlusion. On the contrary, the data captured from roadside cameras has its inherent strengths in terms of robustness to occlusion and long-time event prediction, since they are collected from cameras mounted on poles a few meters above the ground. The comparisons between two different views of data are depicted in Fig.~\ref{fig:different_view}.

The importance of roadside perception is listed as follows: (1) Cooperative to Autonomous driving (AD). AD still faces safety challenges and uncontrolled threats due to blind spots. Instead, the roadside view can cover the blind spots for two extra advantages over car views: 
a long-range global perspective to extend vehicles' perception field spatially and temporally and global trajectory prediction for safety. For example, a pedestrian walking behind a parked vehicle might suddenly crash into a moving vehicle since vehicle sensors fail to detect abrupt changes in the environment owing to the limited perceptual range or heavy occlusion. On the contrary, the roadside view is capable of behavior prediction timely.
(2) Global perception.
Further objects are occluded (even with 360$^{\circ}$ sensors) by closer objects in existing car-view datasets, causing blind spots.
Thanks to roadside cameras mounted overhead, the invisible region is now visible. Besides, Autonomous vehicles (AV) can be informed to choose a faster lane when having a dead car in the queue since the roadside view perceives globally.
(3) Cost-efficient. In terms of cost, it is worthy for ensuring safety by cooperative perception and cost-efficient since information from roadside cameras can broadcast to all surrounding AVs. 
(4) Intelligent traffic control. The roadside perception also facilitates smart traffic control and flow management. The critical contribution of roadside perceptual systems in facilitating a safer and more intelligent autonomous driving system has been acknowledged in many works~\cite{wang2020v2vnet,rauch2012car2x,chen2019cooper}.

However, existing researches on the roadside perceptual ability focus only on 2D tasks such as 2D detection and tracking, the ability of 3D localization is still under-explored\cite{naphade20192019,tan2019multi,naphade20215th}. In this work, we focus on monocular 3D detection that localizes objects in 3D space from a single image. Although abundant perception datasets have been published to fuel the development in autonomous driving from vehicle view, such as KITTI\cite{geiger2012we}, nuScenes\cite{caesar2020nuscenes}, A*3D\cite{pham20203d} and Waymo\cite{sun2020scalability},
none of them is designed particularly for the overlooked roadside 3D perception task.
We hence release the first large-scale high-diversity \textbf{Ro}adside  \textbf{Pe}rception Dataset (Rope3D), with the hope of bridging this gap.
Compared with the existing vehicle view datasets, the roadside perceptual data can be different in three ways. 
First, the ambiguity lies everywhere due to various cameras specifications such as distinct pitch angles of the viewpoint, mounting heights as well as various roadside environments, which increases the difficulty of monocular 3D detection tasks to a great extent. 
Second, since the roadside cameras are mounted on the poles instead of on top of the vehicle in frontal view, thus the assumption of the camera's optical axis being parallel to the ground is no more valid, leading to the incompatibility of directly applying the existing monocular 3D detection approaches using this prior.
Third, due to a much larger sensible range of the roadside perceptual system, a larger number of objects are expected to observe in roadside view, increasing the density and difficulty of a perceptual system. 
All these differences prevent directly applying most existing 3D detection methods. We hence tailor existing monocular 3D object detection methods to the roadside application.

\begin{figure*}[htb]
 \begin{center}
 \vspace{-2mm}
\includegraphics[width=17.3cm]{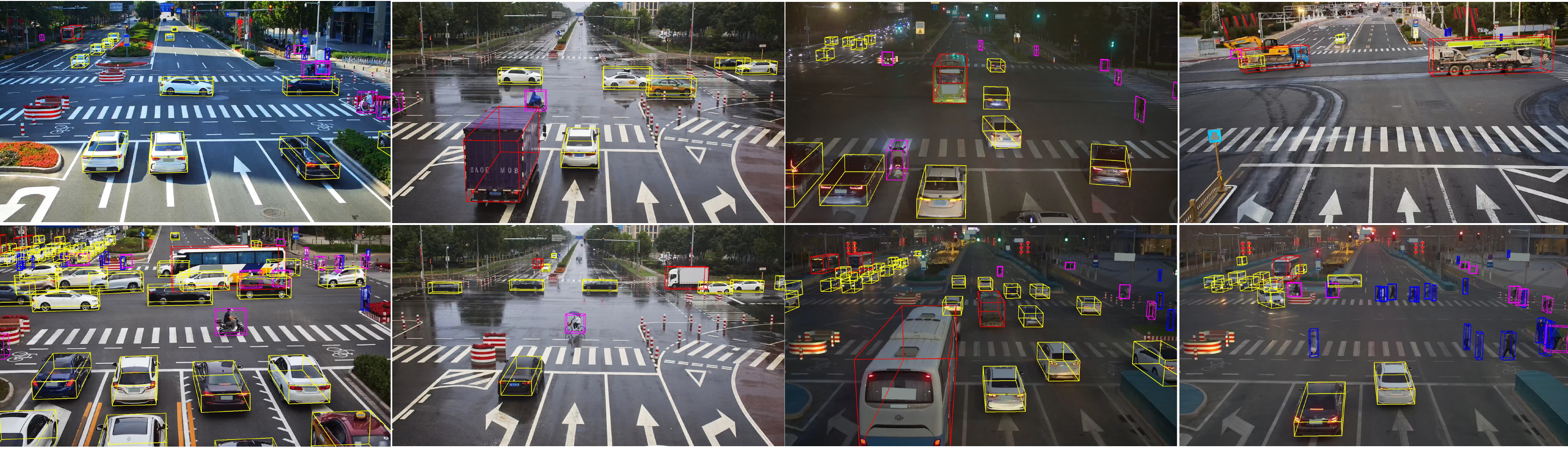}
 \end{center}
 \vspace{-4mm}
 \caption{The examples of collected samples under different weather and conditions. From left to right, each column corresponds to clear/sunny, rainy, night and dawn/dusk. More samples can be found in supplementary material.}
  \vspace{-1mm}
 \label{Figure2}
\end{figure*}

To summarize, our contributions are as follows:
\begin{itemize}
\item{
We present the first challenging high-diversity roadside dataset termed ``Rope3D'', consisting of 50k images and over 1.5M 3D objects collected across a variety of lighting conditions (daytime / night / dusk), different weather conditions (rainy / sunny / cloudy), and distinct road scenes with different camera specifications like focal length and viewpoints.}
\item{We specially tailor current frontal-view monocular 3D detection methods
to deal with the roadside view data and conduct a comprehensive study with the new 3D detection metrics particularly designed for roadside 3D detection tasks, hoping to facilitate the development of monocular 3D perception tasks in roadside scenarios.}
\end{itemize}

\section{Related work}
\label{sec:relatedwork}
\subsection{Frontal-view Autonomous Driving Datasets}
Recent years have witnessed great progress in autonomous driving, thanks to a great number of large-scale traffic scene datasets. ~\cite{gahlert2020cityscapes,xiang2014beyond,choi2018kaist,ma2019trafficpredict,l5dataset,pham20203d,bhattacharyya2021euro} As a pioneer work, KITTI\cite{geiger2012we} provides multimodal data and opens a leader board for a variety of tasks. Although the raw point clouds from LiDAR is 360$^{\circ}$ around the collecting vehicle, the annotation is only conducted within the overlapping frontal view of the camera.
To tackle the limitation, the Honda Research Institute 3D Dataset (H3D)\cite{patil2019h3d} supplies a total of 1.1M 3D boxes in full 360 view, which accelerates full-surround multi-object detection and tracking. 
Another 360$^{\circ}$ view multimodal 3D detection dataset is introduced in nuScenes\cite{caesar2020nuscenes}, providing over 1.4M annotated 3D boxes in 1000 scenes, including nighttime and rainy conditions. 
The ApolloScape and ApolloCar3D\cite{huang2019apolloscape,song2019apollocar3d} focus on the pixel-wise semantic segmentation task, including scene parsing, 3D car instance, lane segmentation tasks. 
The Argoverse dataset\cite{Argoverse19} designs for vehicle perception tasks such as 3D tracking and motion forecasting. 
The Waymo Open Dataset\cite{sun2020scalability} consists of over 1000 scenes and 12M 3D boxes in urban and suburban scenarios, under various weather and lighting conditions. 
Although the above-mentioned datasets have fueled the innovation on autonomous driving, they are all designed for vehicle view perception. However, there is a lack of a dataset helping us to effectively carry out research on 3D localization under roadside surveillance cameras. A related work is from BoxCars~\cite{sochor2018boxcars}, which performs fine-grained vehicle recognition by estimating the projected vertices of 3D bounding box on the image rather than real-world location, size, and orientation. Another contemporary work\cite{dairv2x2022} focuses on the LiDAR-based 3D detection task.
The comparisons between our roadside 3D dataset and the previous AD datasets are listed in Table \ref{tab_3Ddataset_comp}.

\begin{table*}[htb]
 
 \small
 \centering
\resizebox{\linewidth}{!}{
  \begin{tabular}{l|l|l|l|l|l|l|l|l|l|lll|l}
    \specialrule{0.5pt}{0.1pt}{0pt}
   \hline 
   \multirow{2} {*} {View} &
   \multirow{2} {*} {Dataset}  & \ \multirow{2}{0.8cm} {RGB frames}   & \multirow{2}{*} {Scenes} &
   \multirow{2} {0.9cm} {LiDAR channel} & \multirow{2} {0.8cm} {3D Boxes} & \multirow{2} {0.8cm} {2D Boxes} & \multirow{2} {0.8cm} {RGB resolution} & \multirow{2} {*} {Cls} & \multirow{2} {0.5cm} {Year} & \multicolumn{3}{c|}{Diversity} & \multirow{2} {0.7cm} {Range (m)} \\
    &  &  &   &  &  &  &  &  &  & Rain & Night & Dawn & \\
     \hline
    &  KITTI\cite{geiger2012we} & 15k   & 22 & 64 & 80k & 80k   & 1392$\times$512 & 8    & 2013 &     &      &     & 70 \\
    &  Apollo Scape\cite{huang2019apolloscape} & 144k  & /     &64 & 70k   & 0  & 3384$\times$2710 & 8-35 & 2019 &     & $\checkmark$   &     & 420 \\
    & AS lidar\cite{ma2019trafficpredict} & 90k   & /     & 64 & 475k  & 0 & 1920$\times$1080 & 8     & 2019 &      &     &    & 70 \\
    & Lyft Level 5\cite{l5dataset} & 46k   & 366 & 40 & 1.3M  & 0 & 1920$\times$1080 & 9     & 2019 &     &        &   & / \\
    & A2D2\cite{geyer2020a2d2}  & 12k   & /     & 16 & 9k    & 0 & 1928$\times$1208 & 38    & 2019 &     &     &     & 100 \\
    & Argoverse\cite{Argoverse19} & 22k   & 113 & 32 & 993k  & 0 & 1920$\times$1200 & 15    & 2019 & $\checkmark$   & $\checkmark$    &     & 200 \\
    & H3D\cite{patil2019h3d}   & 27.7k & 160 & 64 & 1M    & 0 & 1920$\times$1200 & 8     & 2019 &    &    &     & 100 \\
    & A*3D\cite{pham20203d} & 39k & /  & 64 & 230k  & 0 & 2048$\times$1536 & 7   & 2020 & $\checkmark$   & $\checkmark$    &     & 100 \\
    & CityScapes 3D\cite{gahlert2020cityscapes} & 5k  &1150 & no$^\dag$ & 27k   & 0  & 2048$\times$1024 & 8     & 2020 &    &     &    & 150 \\
    & nuScenes\cite{caesar2020nuscenes} & 1.4M  & 1000 & 32 & 1.4M  & 0 & 1600$\times$900 & 23    & 2020 & $\checkmark$   &$\checkmark$   &       & 75 \\
    & Waymo Open\cite{sun2020scalability} & 230k  &1150 & 64 & 12M   & 9.9M  & 1920$\times$1080 & 4     & 2020 & $\checkmark$   &$\checkmark$      & $\checkmark$   & 75 \\
     \multirow{-11}{*}{Frontal} & ONCE\cite{mao2021one} & 7M  &1M & 40 & 417k   & 0  & 1920$\times$1020 & 5     & 2021 & $\checkmark$   &$\checkmark$     &   & 200 \\
\hline
& BoxCars116k\cite{sochor2018boxcars} & 116k  & 137 & no     & 116k$^\ddagger$ & 0 & $\sim$128$\times$128 & 6 & 2018 &    &    &    & / \\
\multirow{-2}{*}{Roadside} & Ours & 50k   & 26 & 40/300 &1.5M & 670k  & 1920$\times$1080 & 12    & /     &$\checkmark$   & $\checkmark$   &   $\checkmark$   & 200 \\
    \specialrule{0.8pt}{0.1pt}{0pt}
    \end{tabular}%
    }
    \vspace*{-2mm}
    \caption{The comparison of 3D AD datasets. The top and bottom parts indicate the front view and roadside view datasets, respectively.
    LiDAR channel means the beam number of the LiDAR laser. The 2D boxes number denotes those who only have 2D box annotations.`/' denotes unknown information. $(^\dag)$: No lidar sensors to obtain ground truth, use stereo instead. $(^\ddagger)$: For BoxCars116k dataset, only the projected eight corner points instead of the  3D bounding boxes are provided. In other words, the location, dimension, and orientation of 3D bounding boxes are unknown. Besides, only cropped images around 128$\times$128 rather than the full images are provided.}
    \vspace*{-4mm}
 \label{tab_3Ddataset_comp}
\end{table*}

\subsection{Monocular 3D Object Detection}
Though challenging, monocular-based 3D detection is still an attractive solution especially in autonomous driving systems, where the method predicts the 3D bounding boxes from a single image~\cite{mousavian20173d,liu2019deep,li2021monocular,ma2019accurate,atoum2017monocular,chen2016monocular,xu2018multi,ku2019monocular,li2019gs3d}.
Monocular 3D detection methods can be divided into three categories. 
\textbf{(1) Anchor-based}.
Methods in this category exploit a series of predefined 3D bounding box with a location called ``anchor'' and estimate the offset w.r.t the anchor.
M3D-RPN\cite{brazil2019m3d} leverages a 3D region proposal network and the geometric constrains of 2D and 3D perspectives to directly regress the 3D location and size.
Kinematic3D\cite{brazil2020kinematic} further extends M3D-RPN by leveraging 3D kinematics from monocular videos to improve the overall localization.
\textbf{(2) Keypoint-based}.
Many attempts\cite{liu2020smoke,qin2019monogrnet,ma2021delving,zhang2021objects,wang2021fcos3d} have been made to directly regress the keypoints, and then estimate 3D bounding box size and location from the image position of keypoints by optimization e.g., RTM-3D\cite{li2020rtm3d} and MonoGRNet\cite{qin2019monogrnet} .
\textbf{(3) Pseudo-Lidar / depth based}.
Extra depth estimation modules and/or point cloud guidance are employed to alleviate the lack of accurate depth information.
The pioneering work pseudo-LiDAR \cite{you2019pseudo,wang2019pseudo,weng2019monocular} imitates the LiDAR-based methods by utilizing off-the-shelf depth estimators to convert image pixels into pseudo-LiDAR point clouds, and employs LiDAR-based approaches for further detection.
DA-3Ddet\cite{ye2020monocular} adapts the features from unsound image-based pseudo-LiDAR domain to reliable LiDAR domain for guidance to boost the monocular performance.
UrbanNet\cite{carrillo2021urbannet} utilizes the urban 3D map, including driving lanes, elevation, and slope as prior for assisting the 3D detection task.
The existing monocular 3D detection methods are mainly designed for processing vehicle view data. Due to the domain gap and the distribution shift, a question raise naturally is whether these methods still applicable for roadside applications and if not how can we tailor these methods to the new scenario.

\section{The Roadside Perception Dataset}
\subsection{Specifications}
\noindent{\textbf{Sensors Setup.}}
The roadside data collection is conducted by two sets of sensors, one is the roadside cameras mounted on the pole or beside the traffic light; the other is the LiDAR equipped on a parked/driving vehicle to obtain the 3D point clouds of the same scene. 
For sensor synchronization, we adopt the nearest time matching strategy to find the pairs of image and point clouds within 5 milliseconds error.
\begin{itemize}
\vspace{-2mm}
    \item Roadside cameras: RGB with $1920\times1080$ resolution, 30-60Hz capture frequency and 1/1.8'' CMOS sensor.
    \vspace{-2mm}
    \item LiDAR: (1) HESAI Pandar 40P, 40 laser beams, 10/20Hz capture frequency, $\pm$2$cm$ accuracy, $360^{\circ}$ horizontal FOV, -$25^{\circ}$$\sim$+$15^{\circ}$ vertical FOV, $\leq$ 200$m$ range. 
    (2) Jaguar Prime from Innovusion: 300 beams, 6-20 FPS with $\pm$3$cm$ accuracy, $100^{\circ}$ horizontal FOV, $40^{\circ}$ vertical FOV, $\leq$ 280$m$ range.
\vspace{-2mm}
\end{itemize}

\noindent{\textbf{Coordinate Systems and Calibration.}}
There are three coordinate systems used in the dataset: the World Coordinate (\ie, the Universal Transverse Mercator coordinate system (UTM Coord.)), the Camera Coordinate, as well as the LiDAR Coordinate. 

To obtain the reliable ground truth 2D-3D joint annotation, the calibration between different sensors is required.
First, the camera is calibrated to obtain the intrinsics by detecting the chessboard patterns.
Then the Lidar-to-World calibration is conducted by the vehicle localization module to obtain the high definition (HD) map in UTM Coord. For World-to-Camera calibration, we first project the HD map which contains lane and crosswalk endpoints onto the 2D image to obtain the raw transformation. A bundle adjustment refinement is followed to derive the final transformation. Then Lidar-to-Camera transformation can be obtained by simply multiplying Lidar-to-World and World-to-Camera transformations.

 After obtaining the transformation between the three coordinate systems, we can  easily compute the ground equation $G(\alpha, \beta, \gamma, d)$ by fitting the ground points $[x,\!y,\!z]$ to the ground plane in the camera coordinate, with $\alpha x +\beta y+\gamma z+\!d\!=\!0$.

\subsection{Data Collection and Annotation}
\begin{figure*}[t]
 \begin{center}
  \includegraphics[width=\linewidth]{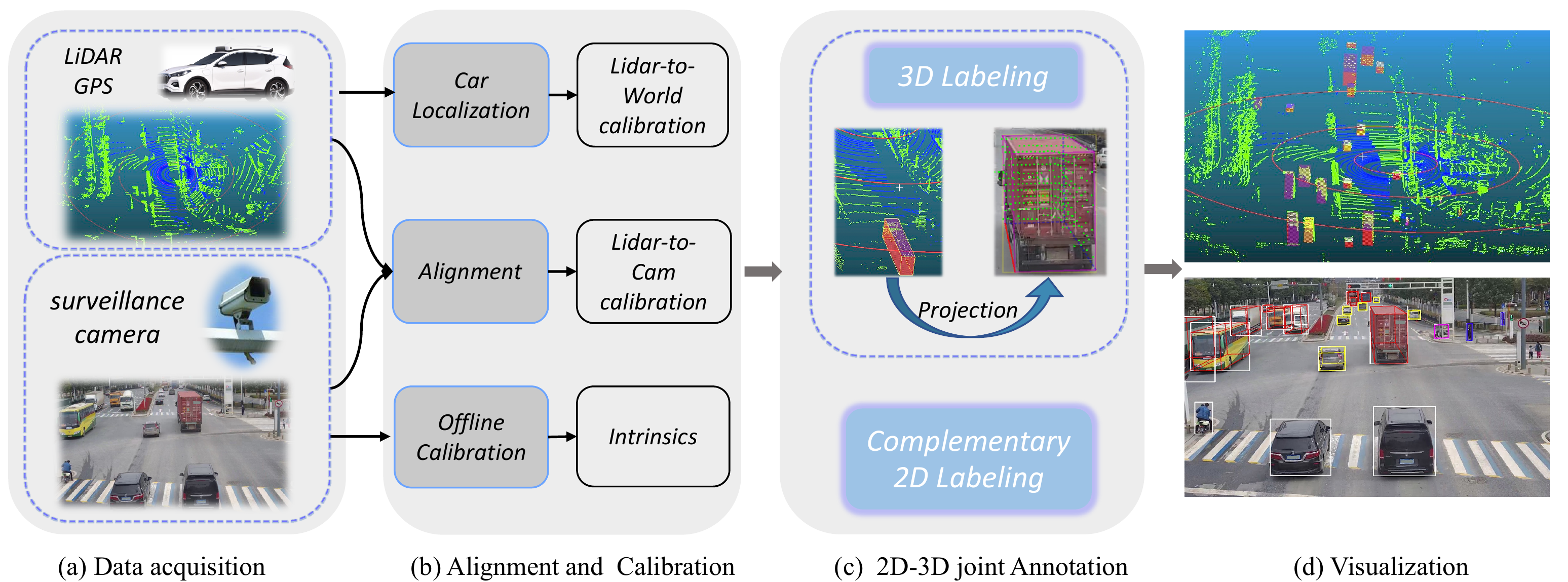}
 \end{center}
 \vspace*{-3mm}
 \caption{The data acquisition and labeling pipeline. Our platform takes the captured roadside images and the point clouds scanned by the LiDAR mounted on a parked/driving vehicle as input. After calibration and alignment between various sensors, the transformations between LiDAR, the world, and the camera are obtained, as well as the ground plane equation and intrinsics. The 2D-3D joint annotation is carried out by projecting the point clouds onto the images and adjusting the 3D bounding boxes manually to fit the 2D instance. For objects that are not scanned by the laser, the 2D complementary labeling is performed on the images only. For example in (d), some objects only have white 2D bounding boxes and no 3D colored annotations due to a lack of 3D points.}
\vspace*{-4mm}
 \label{fig:fig_annotation_pipeline}
 
\end{figure*}

After obtaining the intrinsics as well as the LiDAR-to-Camera transformation, we can collect the 2D-3D data. 
We choose various roadside cameras and let a LiDAR-equipped vehicle park or drive around. 
To keep the high diversity and complexity of the real environment, we collect more than 50k image frames at different times (daytime, night, dawn/dusk), different weather conditions (sunny, cloudy, rainy), different densities (crowded, normal, less traffic), different distributions of traffic elements and so on.
There are totally 13 object classes with their corresponding category, 2D properties (occlusion, truncation)
and the 7-DOF 3D bounding box: Location ($x, y, z$), Size (width-$W$, length-$L$, height-$H$), Orientation (the heading angle-$\theta$).
The full pipeline is illustrated in Fig.~\ref{fig:fig_annotation_pipeline}.
(1)First after obtaining the 3D point clouds and the 2D image (they are within the same space but differ in the viewpoint), we first annotate the 3D bounding boxes directly on the 3D point clouds.
(2) Simultaneously, the annotated 3D bounding boxes will be projected on the 2D image plane, see the top part of Fig.~\ref{fig:fig_annotation_pipeline}(c). We adjust the 3D parameters so that the projected points align with the 2D instance and mainly cover it.
(3) For 2D box annotations, if the objects are scanned by the laser, their 2D box labels in the image are the minimum bounding box of the amodal projections of the eight 3D corners. 
For objects that are heavily occluded or too far to be detected by laser, 2D complementary labeling is conducted to label 2D bounding boxes directly in the image and leave its 3D annotations empty, see the bottom part of Fig.~\ref{fig:fig_annotation_pipeline} (c). 
  
\subsection{Statistics and Analysis}
The collected images have high diversity and inherent ambiguity due to different settings of pitch angles, height, and camera types as well as the manifold scenes.
Two levels of categories are adopted in the dataset. 
The coarse-grained level mainly focuses on the most common traffic elements: Car, Big Vehicle, Pedestrian, and Cyclist. 
To be more fine-grained, Car includes car and van, Big Vehicle can be further divided into truck and bus, and meanwhile, Cyclist can be subdivided into cyclist, motorcyclist, barrow, and tricyclist, since they are driving non-motor vehicles. We have annotated 13 classes, in addition to the above-mentioned categories, there are four extra classes: `traffic cones', `triangle\_plate', `unknown-unmovable', `unknown-movable'. The following statistics are mainly on the coarse and fine-grained classes.

\begin{figure}[t]
 \begin{center}
  \includegraphics[width=\linewidth]{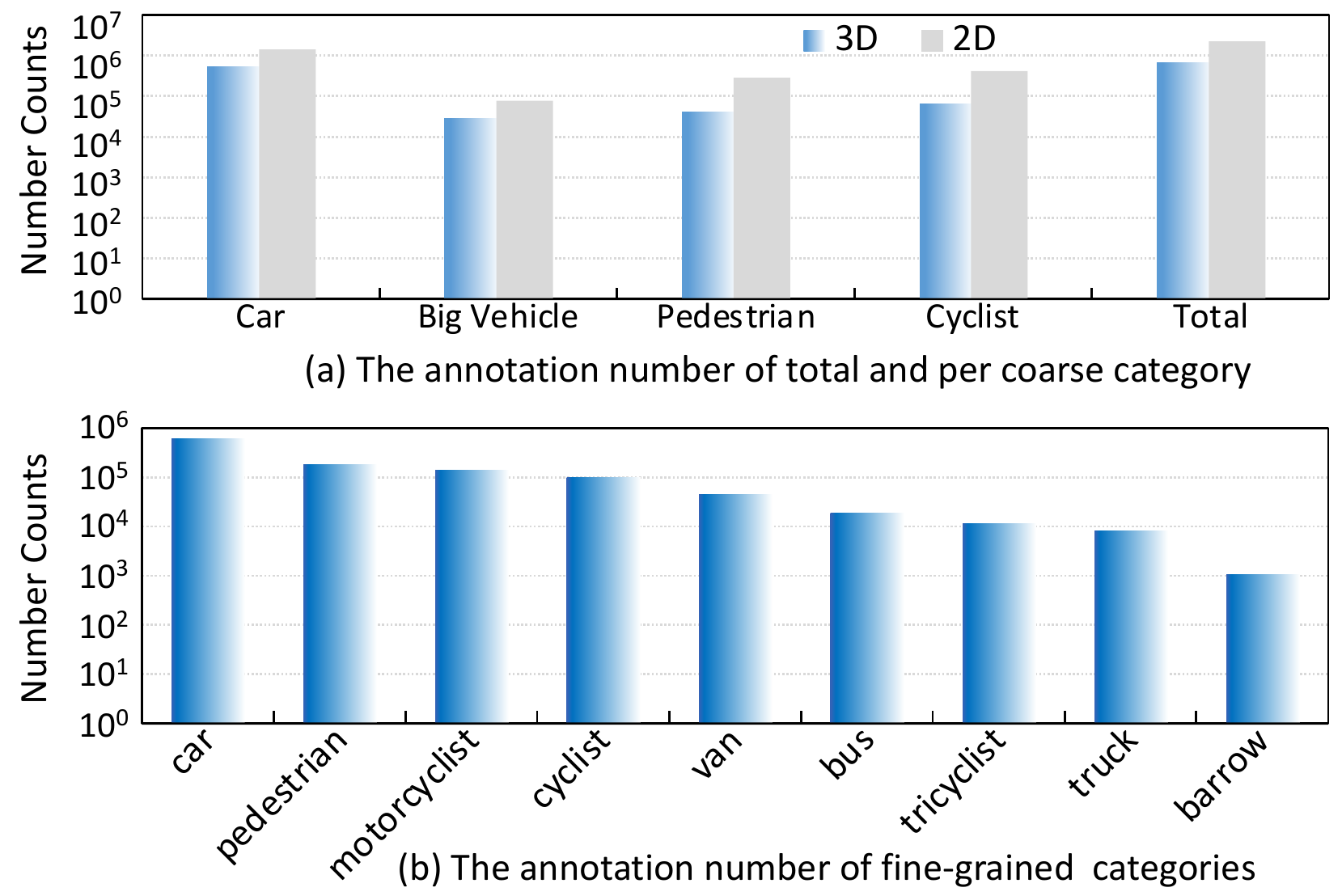}
 \end{center}
 \vspace*{-5mm}
 \caption{The quantities of: (a) coarse level, (b) fine-grained level.
 }
 \label{Figure_number_category}
 \vspace*{-3mm}
\end{figure}

\noindent{\textbf{Quantity distribution.}} We first give an overview analysis of the dataset on the number of 2D and 3D objects in Fig.~\ref{Figure_number_category}. As is stated, 2D objects are more than 3D objects since some objects are not scanned by the LiDAR laser, so they only have 2D annotations. 
We give the detailed number of coarse level and fine-grained levels of the categories, corresponding to (a) and (b) in Fig.~\ref{Figure_number_category}.

\begin{figure}[t]
 \begin{center}
  \includegraphics[width=\linewidth]{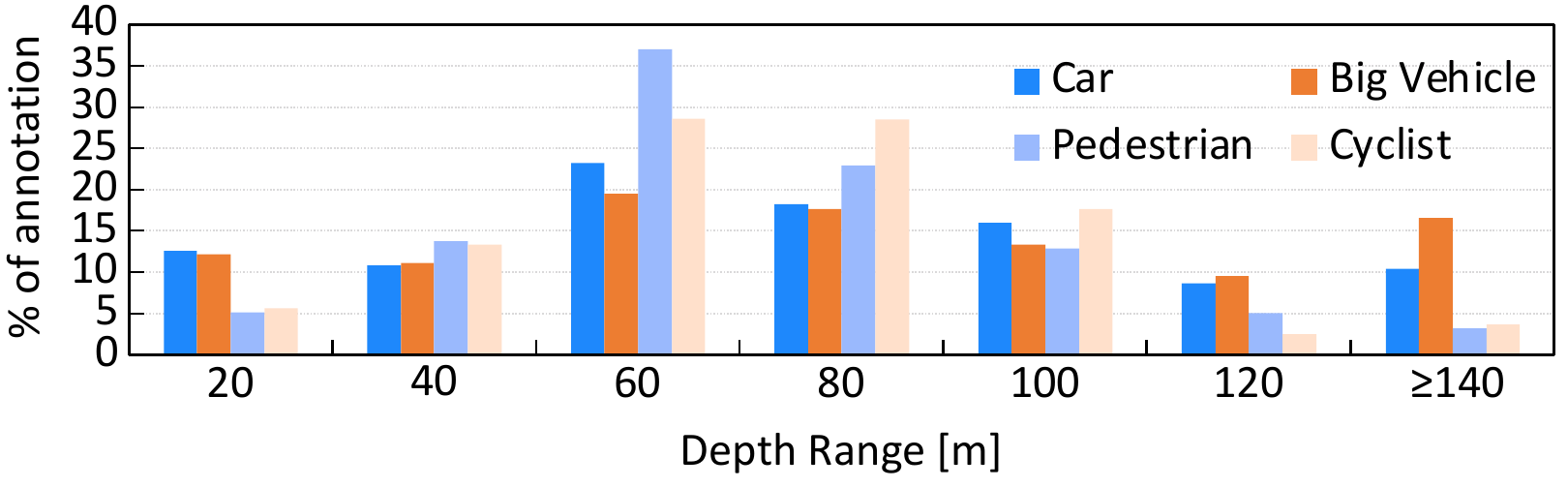}
 \end{center}
  \vspace*{-6mm}
 \caption{The depth distribution of coarse categories.
 }
 \label{Figure_depth_distribution}
  \vspace*{-5mm}
\end{figure}

\noindent{\textbf{Depth distribution.}}
Besides, we analyze the depth distribution of coarse-categories in Fig.~\ref{Figure_depth_distribution}. The depth of the captured 3D objects can range from within 10m to over 140m. Most objects lie between 60 and 80 meters.
 
\begin{figure}[t]
 \begin{center}
  \includegraphics[width=\linewidth]{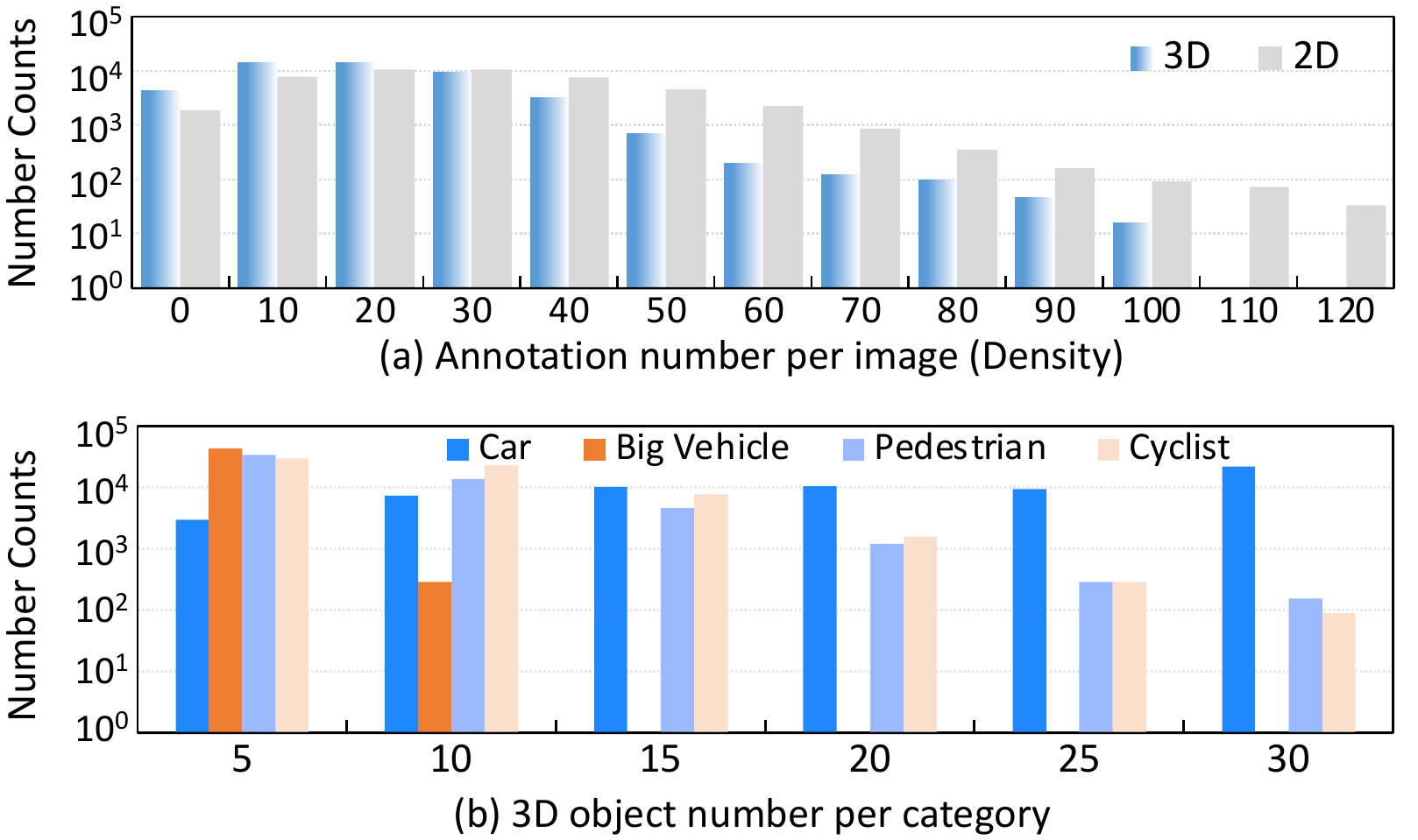}
 \end{center}
 \vspace*{-5mm}
 \caption{Top: The number of total annotated objects per image. Bottom: The number of coarse-category 3D objects per image.}
 \label{Figure_number_density}
 \vspace*{-3mm}
\end{figure}

 \noindent{\textbf{Density.}}
 The density can be a key factor that affects the capability of 3D perception.
 Thus, we analyze the density of the dataset from two aspects in Fig.~\ref{Figure_number_density}. From the global level, we show the 2D and 3D annotated number of each image in (a). The samples can be up to more than one hundred. Compared to other datasets whose densities are KITTI\cite{geiger2012we}: 5.3, nuScenes\cite{caesar2020nuscenes}: 9.7 and A*3D\cite{pham20203d}: 5.9, our dataset has a much higher density (34 and 24 for 2D/3D per image).  
 From the view of coarse categories, the numbers of 3D samples per frame are shown in Fig.~\ref{Figure_number_density} (b). The `Car' category is relatively evenly distributed in densities whereas there are less than 10 big vehicles in each image.
 
\begin{figure}[t]
 \begin{center}
  \includegraphics[width=7.5cm]{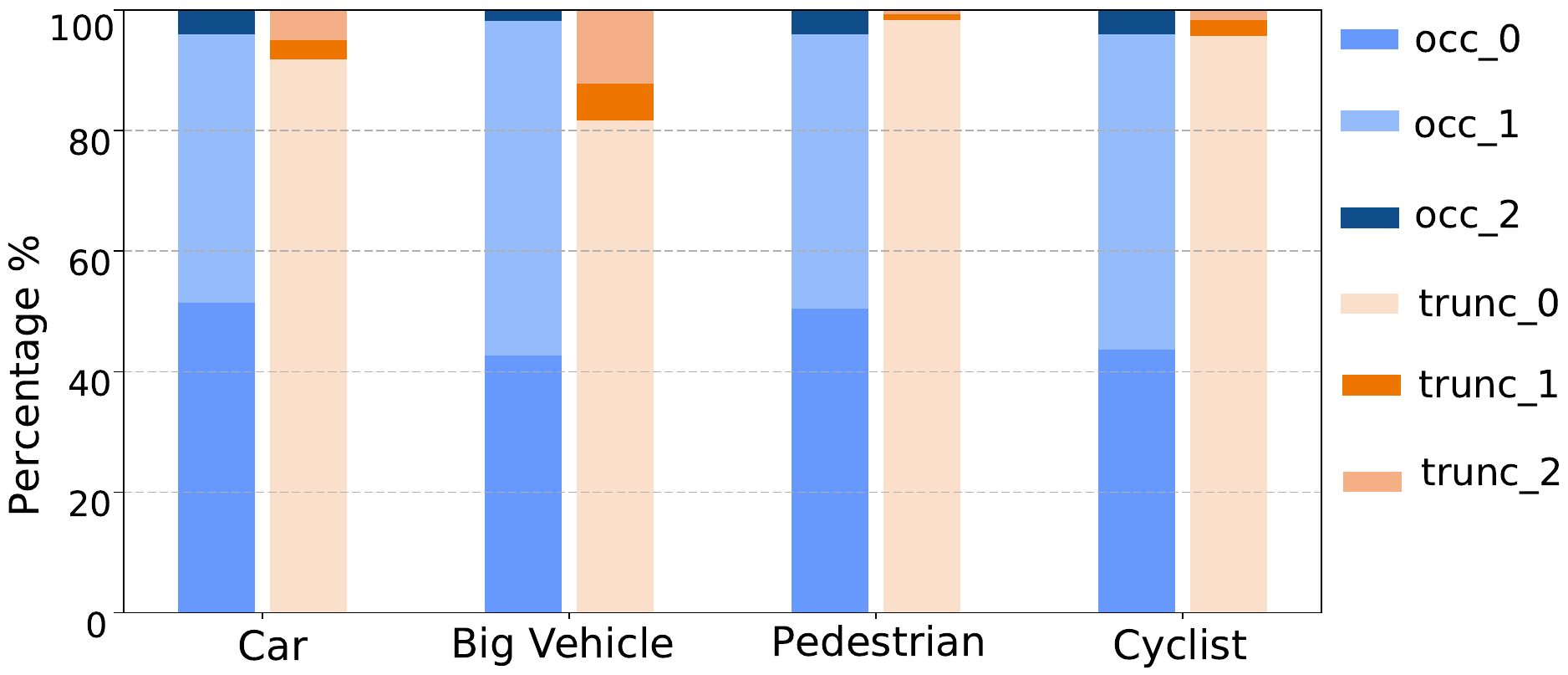}
 \end{center}
 \vspace*{-3mm}
 \caption{The occlusion and truncation distribution of coarse categories. Over half of the objects are partially or heavily occluded.}
 \label{Figure_OccTrunc}
  \vspace*{-5mm}
\end{figure}

\noindent{\textbf{Occlusion and truncation Analysis.}} Next, we annotate three levels for occlusion and truncation attributes. For occlusion, Level 0 denotes no occlusion, 1 and 2 means less / more than 50\% occlusion. For truncation attribute, Level 0 means no truncation, 1 and 2 denote the horizontal and vertical truncation in the image border. The statistics are shown in Fig.~\ref{Figure_OccTrunc}. More than half of the objects are partially or heavily occluded while the occlusion percentage of KITTI is between 5\% to 30\%, which reflects the difficulty of our 3D perception dataset and task.

\begin{figure}[t]
 \begin{center}
  \includegraphics[width=\linewidth]{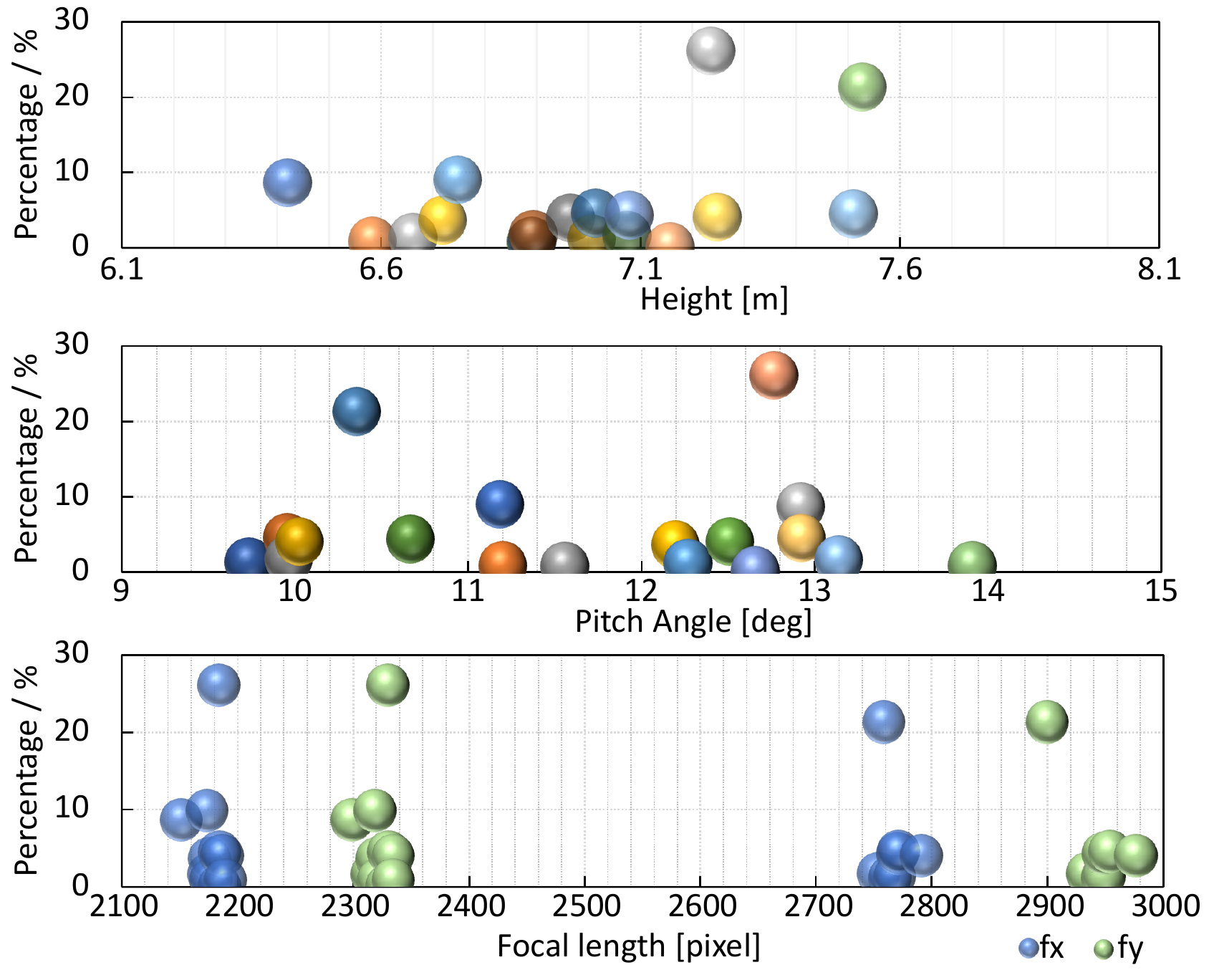}
 \end{center}
 \vspace*{-5mm}
 \caption{The diversity of roadside cameras. From top to bottom are the distribution of mounting heights, pitch angles of the cameras and the focal lengths over the dataset, respectively.}
 \label{Fig_focal_pitch_height}
 \vspace*{-4mm}
\end{figure}

\noindent{\textbf{The ambiguity analysis.}}
The roadside dataset has inherent ambiguity due to the adopted various cameras with different settings of camera specifications, mounting heights, the pitch angles of viewpoint, so on. Thus we analyze the diversity distribution of settings in Fig.~\ref{Fig_focal_pitch_height}.

 

\section{Task and Metrics}
\subsection{The task overview and metrics.}
\noindent{\textbf{Task Overview.}}
The monocular 3D perception task is to localize the objects given ambiguous images captured under various settings and scenes, including different camera specifications, viewpoints, and mounting positions.

\noindent{\textbf{Evaluation Metrics.}}
There are multiple evaluation metrics for AD datasets. In KITTI\cite{geiger2012we}, the 11-point Interpolated Average Precision metric (AP$|_{R_{11}}$) and the average orientation similarity (AOS) are proposed to assess the localization and orientation performance. \cite{simonelli2019disentangling} suggests 40 recall positions instead of 11 positions for a more fair comparison. 
The nuScenes\cite{caesar2020nuscenes} consolidates the mean AP and five True Positive (TP) error types into the nuScenes detection score (NDS), including translation, scale, orientation, velocity and attribute error types. 
Inspired by the AP metric, we adopt AP$|_{R_{40}}$~\cite{simonelli2019disentangling}, \ie,
\begin{equation}
\begin{aligned}
	AP|_{R} =\frac{1}{\left |R\right |} \sum_{r\in R}^{ } \underset{\tilde{r }:\tilde{r }\ge r}{max} \rho \left ( \tilde{r } \right )  
	\label{eq:eq1_AP}
\end{aligned}
\end{equation}
where $\rho\left ( \tilde{r } \right ) $ is the precision at a certain recall theshold $r \in \left \{ 1/40, 2/40, ...,1  \right \}$.  To facilitate a comprehensive evaluation $w.r.t.$ a certain factor such as orientation, we decouple the consolidated metric into several sub-metrics. 

\noindent{\textbf{Average Ground Center Similarity.}}
The distance between objects on the ground plane indicates the risk of collision, we hence compute the ground Euclidean distance by projecting the object center points onto the ground plane:
\begin{equation}
\begin{aligned}
ACS =\frac{1}{\left | D \right |}  \sum_{s \in D}^{} \left ( 1- \min  \left (1,  \frac{\Delta^c_{s}}{C_s}   \right )  \right )
	\label{AGCD}
\end{aligned}
\end{equation}
where $D$ is set of true positive samples, ${C_s}$ is the norm of the GT ground center, $\Delta^c_{s}$ is the Euclidean distance between predicted ground center and GT ground center of sample $s$. $|D|$ is the total number of true positive objects.

\noindent{\textbf{Average Orientation Similarity.}}
The Average Orientation Similarity (AOS) is introduced to measure how well the orientation is estimated, which is defined similarly as \cite{sun2020scalability} by,
\begin{equation}
\begin{aligned}
AOS=\frac{1}{\left | D \right |}  \sum_{s \in D}^{} \frac{1+cos\left (2* \Delta^{\theta}_{s} \right ) }{2}
	\label{AOS}
\end{aligned}
\end{equation}
where $\Delta^{\theta}_{s}$ is the angle difference of sample $s$, and $cos(2*\Delta^{\theta}_{s})$ means that during evaluation, we don't distinguish whether the head or tail of the object is facing the camera.

\noindent{\textbf{Average Area Similarity.}}
We measure the ground occupancy of the prediction w.r.t the GT in term of area, where $\Delta^{A}_{s}$ is the absolute area difference and $A_s$ is the ground truth area. 
\begin{equation}
\begin{aligned}
AAS =\frac{1}{\left | D \right |}  \sum_{s \in D} \left ( 1- \min  \left (1,  \frac{\Delta^{A}_{s} }{A_s}   \right )  \right )
	\label{ASS}
\end{aligned}
\end{equation}


\noindent{\textbf{Average Four Ground Points Distance and Similarity.}}
We also compute the average distance of four ground vertices of the 3D bounding box ($AGD$), since it consolidates the location, orientation, and width/length together. 
\begin{small}
\begin{equation}
\begin{aligned}
AGD =\frac{1}{\left | D \right | }\sum_{s \in D}^{} \left ( \frac{1}{K} \sum_{g=0}^{K-1} \left | s_{g} -\widehat{s_{g} }  \right | \right ) 
\label{AGD4}
\end{aligned}
\end{equation}
\end{small}
where $\widehat{s_{g}}$ and $s_{g}$ are the $g$th predicted and GT ground points of smaple $s$, respectively. $K$ = 4 is the total number of ground points. To be consistent with other similarity metrics, we define the $AGS$ (ground points similarity) as:
\begin{small}
\begin{equation}
\begin{aligned}
AGS \!= \!\frac{1}{\left | D \right | }\!\sum_{s \in D}^{}\!
\left(1 \!-\! \min \!\left (1, \frac{1}{K}\!\sum_{g=0}^{K-1}\!\frac{\left | s_{g} -\widehat{s_{g}} \right | }{\left | \hat{c}  \right |}\right) \right)\!
	\label{AGD4_sim}
\end{aligned}
\end{equation}
\end{small}


Assume $S=(ACS+AOS+AAS+AGS)/4$, we consolidate into $\text{Rope}_\text{score}$ by reweighting the 3D AP and the proposed similarities metrics with $\omega_1=$8 and $\omega_2=$2.
\begin{small}
\begin{equation}
\begin{aligned}
 Rope_{score} \! = (\omega_1 *AP+\omega_2 *S)/(\omega_1 +\omega_2)
	\label{Rope_score} 
\end{aligned}
\end{equation}
\end{small}

\subsection{3D Roadside Perception Task}
As is illustrated in Sec.~\ref{sec:intro}, due to the inherent ambiguity of roadside data caused by diverse camera specifications (various intrinsics and mounting positions, et.al.), the 
existing frontal-view monocular 3D object detection approaches can not be directly applied to the Rope3D Dataset.
Hence we make simple-and-effective attempts to alleviate the ambiguity problem by utilizing camera specifications and encoding the ground knowledge. Two modifications (early-fusion and deep-fusion) to incorporate the depth map of the ground plane with RGB image and two kinds (integrate and multi-gridded) of ground planes are made to alleviate the multi-focal ambiguity.


\noindent{\textbf{Adaptations by leveraging ground planes.}}
 We adopt the ground plane equation $G(\alpha, \beta, \gamma, d)$ and camera intrinsic $K^{3\times3}$ to generate the depth map $D_G$ of the ground plane with the same size as the image.
\begin{small}
\begin{equation}
\begin{aligned}
\left\{\begin{matrix}
Z\begin{bmatrix} x, y, 1 \end{bmatrix}^T = K^{3\times3}\begin{bmatrix} X, Y, Z \end{bmatrix}^T\\
G^{1\times4} \begin{bmatrix} X, Y, Z, 1 \end{bmatrix}^T = 0
\end{matrix}\right.
\end{aligned}
\end{equation}
\end{small}
where $[x, y]$ is the pixel in the image coordinates, $[X,Y,Z]$ is the corresponding 3d point in camera coordinate that lies on the ground plane. 
Thus the depth $Z$ can be derived with the known 2d image points and the ground plane equation $G$. 
We incorporate the ground depth map with the RGB appearance feature by early fusion and deep fusion. The first one is directly concatenating the depth map with the original RGB channels as input, and the second is adopting
another siamese network for depth feature extraction
and further weighted fusion of the two depth predictions.
The performances of these two methods are similar and we
hence only report the results by concatenation on the anchor-based M3D-RPN and the keypoint-based MonoDLE
and MonoFlex approaches. We believe more sophisticate
approaches might further improve the performance,
which is out of the scope of this paper.
In addition, two different formats of ground planes are attempted. One is to fit the entire ground within
the visual field to a single plane, which is represented by
the ground plane equation. Another is to divide the entire
ground into multiple small grids, and each grid is represented by a ground equation.

\section{Experiments}
\begin{table*}[t]
\begin{center}
  \resizebox{\linewidth}{!}{  
\begin{tabular}{c| c | c | c | cc cc |cc cc }
\specialrule{0.8pt}{0.1pt}{0pt}
\multirow{3}{*}{Setting} & \multirow{3}{*}{Method} & \multirow{3}{*}{Backbone} &\multirow{3}{*}{Branch} & \multicolumn{4}{c|}{IoU = 0.5} & \multicolumn{4}{c}{IoU = 0.7} \\ 
&  &  &   & \multicolumn{2}{c}{Car} & \multicolumn{2}{c|}{Big Vehicle} & \multicolumn{2}{c}{Car} & \multicolumn{2}{c}{Big Vehicle} \\ 
&  &  &   & AP$_{\text{3D}{|\text{R40}}}$ & Rope$_\text{score}$ & AP$_{\text{3D}{|\text{R40}}}$ & Rope$_\text{score}$  & AP$_{\text{3D}{|\text{R40}}}$ & Rope$_\text{score}$ & AP$_{\text{3D}{|\text{R40}}}$ & Rope$_\text{score}$\\ 
\hline
\multirow{5}{*}{$\mathbb{I}$} 	
& M3D-RPN-$\bm{(G)}$~\cite{brazil2019m3d} & \multirow{1}[0]{*}{ResNet34}  & A 
&54.19 & \multicolumn{1}{c|}{62.65}	&33.05 &  44.94 
&16.75 & \multicolumn{1}{c|}{32.90} &6.86  &  24.19 \\

&  M3D-RPN-$\bm{(D)}$~\cite{brazil2019m3d} & \multirow{1}[0]{*}{ResNet34} 	& A 
& 67.17 & \multicolumn{1}{c|}{73.14} & 39.06 & 49.95 
& 33.94 &\multicolumn{1}{c|}{46.45}  & 11.28 &  28.12\\

&  Kinematic3D-$\bm{(G)}$~\cite{brazil2020kinematic} & \multirow{1}[0]{*}{DenseNet121} & A	 
&50.57  & \multicolumn{1}{c|}{58.86}&	37.60&  48.08 
&17.74  & \multicolumn{1}{c|}{32.99}&   6.10&   22.88\\
 &  MonoDLE-$\bm{(G)}$~\cite{ma2021delving} & \multirow{1}[0]{*}{DLA-34} & K  
 & 51.70 &\multicolumn{1}{c|}{60.36} & 40.34 & 50.07  
 & 13.58 & \multicolumn{1}{c|}{29.46}&9.63 &25.80\\
&  MonoDLE-$\bm{(D)}$~\cite{ma2021delving} & \multirow{1}[0]{*}{DLA-34} & K  
 & 77.50 &\multicolumn{1}{c|}{80.84} & 49.07 & 57.22  
 & 54.53 & \multicolumn{1}{c|}{62.48}&17.25 &32.00\\

 & MonoFlex-$\bm{(G)}$~\cite{zhang2021objects} & \multirow{1}[0]{*}{DLA-34}  & K	 & 60.33 & \multicolumn{1}{c|}{66.86}&	37.33 &47.96  
 & 33.78 &\multicolumn{1}{c|}{46.12} &   10.08 &26.16\\
& MonoFlex-$\bm{(D)}$~\cite{zhang2021objects} & \multirow{1}[0]{*}{DLA-34}  & K	 & 59.78 & \multicolumn{1}{c|}{66.66}&	59.81 &66.07 
 & 35.64 &\multicolumn{1}{c|}{47.43} &   24.61 &38.01\\
\hline
\hline
\multirow{5}{*}{$\mathbb{II}$} 	
& M3D-RPN-$\bm{(G)}$~\cite{brazil2019m3d} & \multirow{1}[0]{*}{ResNet34}  & A  
& 21.75& \multicolumn{1}{c|}{36.40}	&21.49 & 35.49 
& 6.05&\multicolumn{1}{c|}{23.84} &2.78 & 20.82  \\

&  M3D-RPN-$\bm{(D)}$~\cite{brazil2019m3d} & \multirow{1}[0]{*}{ResNet34} 	& A &36.33 & \multicolumn{1}{c|}{48.16} & 24.39 & 37.81
&11.09 &\multicolumn{1}{c|}{28.17}  &3.39 & 21.01\\

&  Kinematic3D-$\bm{(G)}$~\cite{brazil2020kinematic} & \multirow{1}[0]{*}{DenseNet121} & A	 
& 23.56 & \multicolumn{1}{c|}{37.05}&13.85	&28.58   
&5.82  & \multicolumn{1}{c|}{23.06}&	1.27& 18.92\\
 &  MonoDLE-$\bm{(G)}$~\cite{ma2021delving} & \multirow{1}[0]{*}{DLA-34} & K  
 & 19.08&\multicolumn{1}{c|}{33.72} & 19.76 & 33.07  
 & 3.77 & \multicolumn{1}{c|}{21.42}&2.31 &19.55\\
&  MonoDLE-$\bm{(D)}$~\cite{ma2021delving} & \multirow{1}[0]{*}{DLA-34} & K  
 & 31.33&\multicolumn{1}{c|}{43.68} & 23.81 & 36.21
 & 12.16 & \multicolumn{1}{c|}{28.39}&3.02 &19.96 \\
 
 & MonoFlex-$\bm{(G)}$~\cite{zhang2021objects} & \multirow{1}[0]{*}{DLA-34}  & K	 & 32.01 & \multicolumn{1}{c|}{44.37}&	13.86 & 28.47
 & 10.86 &\multicolumn{1}{c|}{27.39} &0.97 &18.18\\
& MonoFlex-$\bm{(D)}$~\cite{zhang2021objects} & \multirow{1}[0]{*}{DLA-34}  & K	 & 37.27 & \multicolumn{1}{c|}{48.58}&	47.52 & 55.86
 & 11.24 &\multicolumn{1}{c|}{27.79} &13.10 &28.22\\

\specialrule{0.7pt}{0.1pt}{0pt}
\end{tabular}
}
\end{center}
\caption{Overall performance of the monocular 3D object detection approaches on the Rope3D Dataset with IoU = 0.5 and 0.7 under two train-val splitting settings: the homologous ($\mathbb{I}$) and the heterologous ($\mathbb{II}$). -$\bm{(G)}$ denotes adapting the ground plane, -$\bm{(D)}$ means using the depth map of ground. The abbr. in the branch column denotes: A: anchor-based, K: keypoint-based.}
\label{tab_performance_overall}
\end{table*}

\begin{table*}[t]
\begin{center}
  \resizebox{\textwidth}{!}{  
\begin{tabular}{l | c | c | c c c  c c c c c c }
\specialrule{0.8pt}{0.1pt}{0pt}
\multirow{2}{*}{Setting} & \multirow{2}{*}{Method} & \multirow{2}{*}{Backbone} & \multicolumn{8}{c}{AP$_{\text{3D}{|\text{R40}}}$[Mod] / Rope$_\text{score}$ } &  \\ 
& & & car & van & bus & truck & cyclist & motorcyclist & tricyclist & pedestrian  \\ 
\hline
\multirow{3}{*}{$\mathbb{I}$} & 
\multirow{1}{*}{KM3D~\cite{li2021monocular} } 		&  ResNet34	& 8.97 / 25.09	& 7.77 / 23.79	& \textbf{8.07} / 23.89	& 4.94 / 20.59	&1.81 / 17.34 	&3.61 / 19.34	& 14.39 / 27.85	& 0.37 / 17.93 	\\ 
&
\multirow{1}{*}{KM3D-$\bm{(G)}$~\cite{li2021monocular} } 	&  ResNet34	& 9.83 / 26.60	& 13.16 / 29.26	& 4.19 / 22.05	& \textbf{18.42} / \textbf{32.40}	& 11.35 / 27.24	& 11.45 / 27.24	& 19.13 / 33.50	& 9.90 / 26.28 	\\ 
&
\multirow{1}{*}{KM3D-$\bm{(GG)}$~\cite{li2021monocular} } 	& ResNet34 	& \textbf{9.86} /\textbf{26.64} & \textbf{15.71} / \textbf{31.30} & 7.66 / \textbf{24.48}	& 12.67 / 27.71	& \textbf{13.23} / \textbf{28.94}	& \textbf{15.08} / \textbf{30.14}	& \textbf{19.97} / \textbf{34.13}	& \textbf{11.92} / \textbf{27.90} 	\\ 
\hline
\hline
\multirow{3}{*}{$\mathbb{II}$} & 
\multirow{1}{*}{KM3D~\cite{li2021monocular} } 		&  ResNet34	& 5.89 / 22.91 & 2.91 / 20.26	& 21.20 / 34.30	& 25.86 / 37.46	& 1.36 / 17.14 & 4.67 / 20.03	& 2.40 / 19.48	& 0.29 / 17.93  	\\ 
&
\multirow{1}{*}{KM3D-$\bm{(G)}$~\cite{li2021monocular} } 	&  ResNet34	& 17.39 / 32.71	& 30.48 / 43.22	& \textbf{21.25} / \textbf{35.20}	& 34.93 / 45.84	& 24.98 / 38.30	& 14.49 / 29.47	& 47.47 / 56.79	& 12.61 / 28.44  	\\ 
&
\multirow{1}{*}{KM3D-$\bm{(GG)}$~\cite{li2021monocular} } 	&  ResNet34	& \textbf{23.70} / \textbf{37.90}	& \textbf{31.37} / \textbf{44.04}	& 19.99 / 34.61	& \textbf{37.65} / \textbf{48.34}	& \textbf{26.38} / \textbf{39.58}	& \textbf{16.58} / \textbf{30.82}	& \textbf{54.03} / \textbf{62.46}	& \textbf{12.81} / \textbf{28.46}	\\ 

\specialrule{0.8pt}{0.1pt}{0pt}
\end{tabular}
}
\end{center}
\caption{Performance of KM3D with different ground plane formats on the Rope3D Dataset under two train-val splitting settings: the homologous ($\mathbb{I}$) and the heterologous ($\mathbb{II}$). -$\bm{(G)}$ denotes adapting the ground plane equation, -$\bm{(GG)}$ means using the gridded ground planes. IoU = 0.25 for non-motor vehicles and pedestrian, IoU = 0.5 for motor vehicles.}
\label{performance_of_groundplane}
\end{table*}

\begin{table}[t]
\begin{center}
\resizebox{\linewidth}{!}{
\begin{tabular}{l | c | c c c c  }
\specialrule{0.8pt}{0.1pt}{0pt}
\multirow{2}{*}{Method} & {Range} & \multicolumn{4}{c}{AP$_{\text{3D}{|\text{R40}}}$ / Rope$_\text{score}$ }   \\ 
& (m) & Car & Big Vehicle  & Cyclist & Pedestrian  \\ 
\hline
\multirow{5}{*}{MonoDLE-$\bm{(G)}$} 	
	& all 	 & 19.08 / 33.72 & 19.76 /	33.07 & 10.93 /	26.44 & 3.72 /	21.42 	\\ 
	& 0-30 	 & 31.43 /	43.12 & 37.36 /	46.69 & 19.83 /	33.28 & 10.26 /	26.61 	\\ 
	& 30-60  & 10.42 /	26.68 & 8.68 /	24.02 & 8.11 /	24.23 & 3.90 /	21.66  	\\ 
	& 60-90  & 18.60 /   33.42 & 32.46 / 44.15 & 9.29 /	25.23 & 2.37 /	20.28  	\\ 
	& 90-120 & 11.84 /	28.05 & 10.29 /	26.09 & 9.84 /	25.11 & 2.88 /	20.58 	\\ 
\hline
\multirow{5}{*}{MonoFlex-$\bm{(G)}$} 	
	& all 	 & 32.01 / 44.37 & 13.86 /	28.47 & 44.27 / 53.58 & 25.48 /	39.04 	\\ 
	& 0-30 	 & 15.49 / 30.49 & 27.68 /	39.16 & 61.94 /	67.37 & 37.25 /	48.50 	\\ 
	& 30-60  & 45.69 / 55.33 & 12.18 /	27.48 & 50.70 /	58.78 & 35.74 /	47.37  	\\ 
	& 60-90  & 46.72 / 56.41 & 19.34 /	33.45 & 30.65 /	42.82 & 8.94 / 25.65  	\\ 
	& 90-120 & 14.19 / 30.15 & 1.30 / 18.82 & 9.43 / 25.20 & 4.79 / 22.35 	\\ 
\hline
\multirow{5}{*}{M3D-RPN-$\bm{(D)}$} 	
	& all    & 36.33 / 48.16 & 24.39 / 37.81  & 11.22 /	27.54  & 3.93 /	21.54 	\\ 
	& 0-30 	 & 52.07 / 60.60 & 24.07 / 37.18  & 16.77 / 31.80  & 5.19 /	22.53 	\\ 
	& 30-60  & 33.57 / 46.30 & 25.22 / 38.84  & 14.38 /	30.36  & 8.09 /	24.99  	\\ 
	& 60-90  & 24.07 / 38.60 & 39.17 / 50.42  & 5.80 / 23.02  & 1.10 / 19.16 	\\ 
	& 90-120 & 11.19 / 28.23 & 6.55 / 24.06  & 4.23 / 21.48  & 0.13 / 18.02 	\\ 
\specialrule{0.7pt}{0.1pt}{0pt}
\end{tabular}
}
\end{center}
\caption{The performance within different ranges on the heterologous ($\mathbb{II}$) set. We set IoU = 0.5 for Car / Big Vehicle and IoU = 0.25 for Cyclist and Pedestrian, following KITTI's hard-modality. 
}
\label{tab_ranges}
\end{table}
\subsection{Experimental Setup}
Our roadside 3D perception dataset contains 50k images, with the training and validation ratio set to 8:2. We offer two kinds of splitting the training and validation set, $\mathbb{I}$: Homologous, for each scene we select 70\% images and combine them for training, and leave all the rest images for validation.
$\mathbb{II}$: Heterologous, we select 80\% of the cameras with the collected images for training and leave the remaining unseen 20\% (different camera specifications) for validation, which can be used for validating the generalization ability of the monocular 3D object detection approaches.

\noindent{\textbf{Implementation Detail.}}
(1) For M3D-RPN\cite{brazil2019m3d}, we experiment on vanilla and improved approaches with ResNet34\cite{he2016deep} backbone, 
(2) Kinematic3D\cite{brazil2020kinematic} is a monocular video-based 3D object detector with DenseNet121~\cite{huang2017densely} backbone, we only implement the first phase without video knowledge. 
(3) MonoDLE\cite{ma2021delving} is based on the anchor-free one stage detector CenterNet\cite{zhou2019objects} with backbone DLA34~\cite{yu2018deep}. 
(4) MonoFlex\cite{zhang2021objects} is a keypoint based method with modified DLA34~\cite{yu2018deep} backbone.
The training image resolutions are adjusted to fit our dataset.

For objects that contain only 2D annotations without 3D annotations, we compute only losses on 2D attributes. For those objects having 3D labeling, the training loss weights of the 2D and 3D are both set to 1.

\subsection{Main Results and Analysis}
\noindent{\textbf{Performance of vanilla and improved approaches.}}
The performances of monocular 3D detection approaches on the Rope3D Dataset are depicted in Table~\ref{tab_performance_overall}. 
Approaches with suffix $\bm{(G)}$ denotes we customize the corresponding approach with ground plane function to reconnect the 3D locations and 2D projected points even when the optical axis of cameras is not parallel to the ground plane because of the pitch angle.
The improved approach is noted with suffix $\bm{(D)}$. We adopt the 3D $AP|_{R40}$ at the moderate level as well as the proposed metrics for evaluation. 
We find that most approaches have an obvious performance decline from homologous to the heterologous validation set. However, the performance drop is relatively insignificant when applying a 3D detection model trained on the vehicle-view ONCE dataset to nuScenes dataset\cite{mao2021one}. This phenomenon indicates the domain gap caused by various camera specifications and setting positions can not be ignored,
which might be a distinguishable difference between vehicle-view and roadside view applications and should be taken carefully.
By leveraging the depth map of ground planes, we observe an obvious improvement in most methods, even on the heterologous set where training and validating images have different camera specifications. 

\noindent{\textbf{Performance of different ground plane formats.}}
We further analyze the performance by adopting two different formats of ground planes, \ie, by fitting the entire ground within the visual field to a single plane or by dividing the  entire ground into multiple 5m$\times$5m grids piecewisely. 
We carry out the experiment on KM3D\cite{li2021monocular}, a method that predicts 2D keypoints and solves 3D position by minimizing the re-projection error. In other words, it relies on the differentiable geometric constraint to recover 3D location rather than direct prediction, which heavily depends on the accuracy of ground plane. 
As is shown in Table~\ref{performance_of_groundplane}, KM3D-$\bm{(GG)}$, which takes advantage of gridded ground planes, shows better performance on most fine-grained categories. The main reason might be that the piecewise gridded planes better fit the actual the ground plane.

\noindent{\textbf{Performance of different ranges.}}
We further analyze the performance of the models within different ranges from 0 to 120m. 
As is shown in Table~\ref{tab_ranges}, with the depth range increases, most of the performances decrease, especially for 90-120 meters. The reason is owing to two aspects: too small area in the image to extract strong features for learning and much less 3D annotations in far-away regions due to occlusion. MonoFlex-$\bm{(G)}$ shows better performance on cyclists and pedestrians whereas inferior in motor vehicles.



\section{Conclusion}
We propose the first high-diversity challenging roadside monocular 3D perception dataset - Rope3D.
Rope3D is collected from the roadside view with joint 2D-3D annotations, making it unique from any previously released datasets and is particularly designed for the roadside 3D perception.
Furthermore, we specially tailor the existing monocular 3D object detection approaches to the novel dataset, due to its unique viewpoint and inherent ambiguity lying in the various camera specifications and diverse road scenes. We hope to raise the attention to the special view - roadside view, so as to facilitate a safer and more intelligent autonomous driving system.

\noindent{\textbf{Ethical concerns and Limitation.}}
To prevent from being utilized for illegal surveillance,  all the images in the dataset are time-discrete and not allowed for tracking tasks. Note that all the sensitive information including license plates, human faces, names of bus stops, roads, and buildings are totally masked. The attempts for adapting the concurrent vehicle-view 3D detection approaches need further study.

\newpage

\appendix
\section{Additional Dataset Analysis}
\label{sec:Additional_Dataset_Analysis}
\subsection{Detailed Dataset Analisis}%
\noindent{\textbf{Comparison in the object density against other datasets.}} We further compare the average number of 3D annotations per frame of different datasets. As is demonstrated in Table~\ref{tab_detailednumber_comparison}, we compute the density for A*3D Dataset and borrow the statistics from CityScapes 3D\cite{gahlert2020cityscapes} for the following datasets: KITTI, ApolloScapes, Argoverse, nuScenes, Waymo, and CityScapes 3D. Compared with other datasets, we have a high object density across all classes.

\begin{table}[thb!]
\begin{center}
  \resizebox{\linewidth}{!}{  
\begin{tabular}{l | c c c c c}
\specialrule{0.8pt}{0.1pt}{0pt}
 & Car & Big Vehicle & Cyclist & Pedestrian & All \\
\hline
KITTI\cite{geiger2012we} & 4.2 & 0.2 & 0.0 & 0.0 & 4.40 \\
ApolloScapes\cite{huang2019apolloscape} & 11.6 & 0.0 & 0.0 & 0.0 & 11.60\\
Argoverse\cite{Argoverse19} & 4.1 & 0.3 & 0.1 & 0.001  & 4.50\\
nuScenes\cite{caesar2020nuscenes} & 3.0 & 0.6 & 0.07 & 0.07 & 3.74 \\
Waymo\cite{sun2020scalability} & 3.2 & 0.0 & 0.04 & 0.0 & 3.24 \\
CityScapes 3D\cite{gahlert2020cityscapes} & 6.4 & 0.2 & 1.2 & 0.2  & 8.0\\
A*3D\cite{pham20203d} & 3.9 & \textbf{1.0} & 0.3 & 0.6 & 5.8\\
Ours & \textbf{14.0} & 0.6 & \textbf{3.9} & \textbf{5.5 } & \textbf{24.0}\\
\specialrule{0.8pt}{0.1pt}{0pt}
\end{tabular}
}
\end{center}
\caption{The average number of 3D annotations in each image at coarse-level. Compared with other datasets, we have a high object density across all classes.}
\label{tab_detailednumber_comparison}
\end{table}

\begin{figure}[h]
 \begin{center}
  \includegraphics[width=0.85\linewidth]{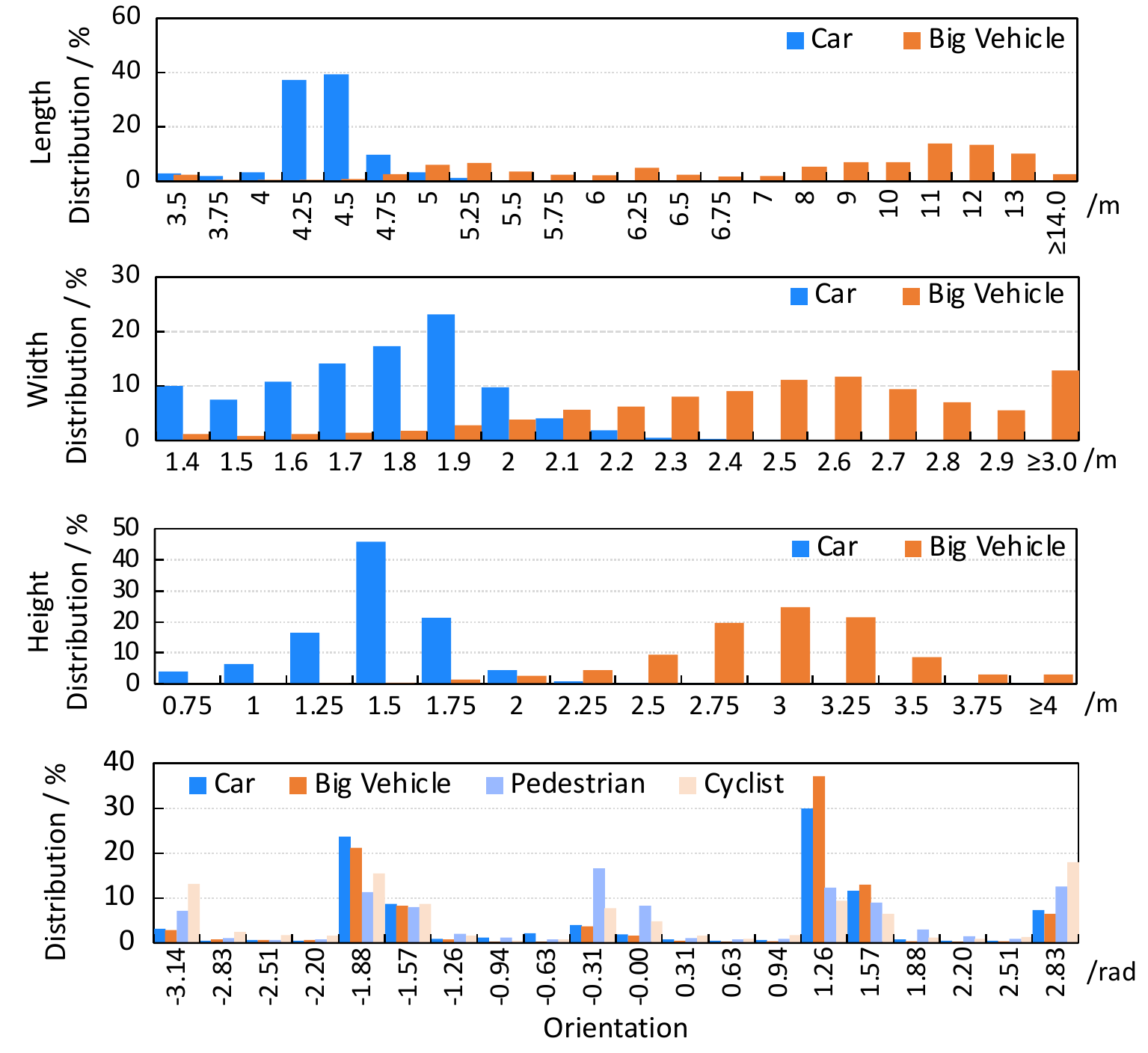}
 \end{center}
 \caption{From top to bottom are the distribution of length, width, height and orientation over the motor vehicles, respectively. Only the 3D sizes of car and big vehicles are counted, since the sizes of pedestrian and cyclists only have little changes.}
 \label{Figure_LHW_ry}
\end{figure}

\noindent{\textbf{Size and orientation.}}
 Only motor vehicles are taken into account for size analysis, \ie, cars, and big vehicles since non-motor categories usually have similar sizes. The size and orientation distributions are presented in Fig.~\ref{Figure_LHW_ry}. Due to various camera specifications and diverse scenes, the high-frequency orientations are not constrained to a single peak.

\noindent{\textbf{The Mean and Std Dev of fine-grained categories.}}
We further compute the mean and standard deviation (Std Dev) of each fine-grained category, which is presented in Table~\ref{tab_LHW_meanstd}. The mean and Std Dev values can be utilized for pre-defining the mean size and the disturbance range. For example, Monoflex \cite{zhang2021objects} estimates the offset of length, width, and height w.r.t the mean values, instead of directly regressing the sizes, which improves the robustness and accuracy of size prediction.
\begin{table}
\begin{center}
\small
  \resizebox{\linewidth}{!}{  
\begin{tabular}{l | c| c  c c}
\specialrule{0.8pt}{0.1pt}{0pt}
 Category & Metric &  Length / m & Height / m & Width /   m \\
 \hline
 \multirow{2}[0]{*}{Car} & mean & 4.247 & 1.325& 1.706  \\
 &                      Std Dev & 0.315 & 0.258 & 0.234 \\
 \hline
  \multirow{2}[0]{*}{Truck} & mean & 7.122 & 2.623& 1.706  \\
 &                      Std Dev & 2.067 & 0.628 & 0.492\\
 \hline
  \multirow{2}[0]{*}{Van} & mean &4.651 &1.750 & 1.757   \\
 &                      Std Dev & 0.429& 0.311& 0.268 \\
 \hline
   \multirow{2}[0]{*}{Bus} & mean & 10.575 & 3.009& 2.533  \\
 &                      Std Dev & 1.806& 0.404& 0.426\\
 \hline
\multirow{2}[0]{*}{Pedestrian} & mean &0.478 & 1.610& 0.501  \\
 &                      Std Dev &0.178 & 0.160& 0.143 \\
 \hline
\multirow{2}[0]{*}{Cyclist} & mean & 1.525 & 1.382 & 0.505  \\
 &                      Std Dev & 0.264 & 0.280 & 0.217 \\
 \hline
 \multirow{2}[0]{*}{Tricyclist} & mean & 2.631 & 1.539 & 1.077  \\
 &                      Std Dev & 0.497 & 0.196 & 0.292 \\
 \hline
\multirow{2}[0]{*}{Motorcyclist} & mean & 1.692&  1.418& 0.613  \\
 &                      Std Dev &0.276 & 0.175& 0.211\\
\specialrule{0.7pt}{0.1pt}{0pt}
\end{tabular}
}
\end{center}
\caption{The mean and Std Dev size of each fine-grained category over the Rope3D Dataset.}
\label{tab_LHW_meanstd}
\end{table}


\subsection{More samples of the Rope3D Dataset.}%
\begin{figure*}[h]
 \begin{center}
  \includegraphics[width=\linewidth]{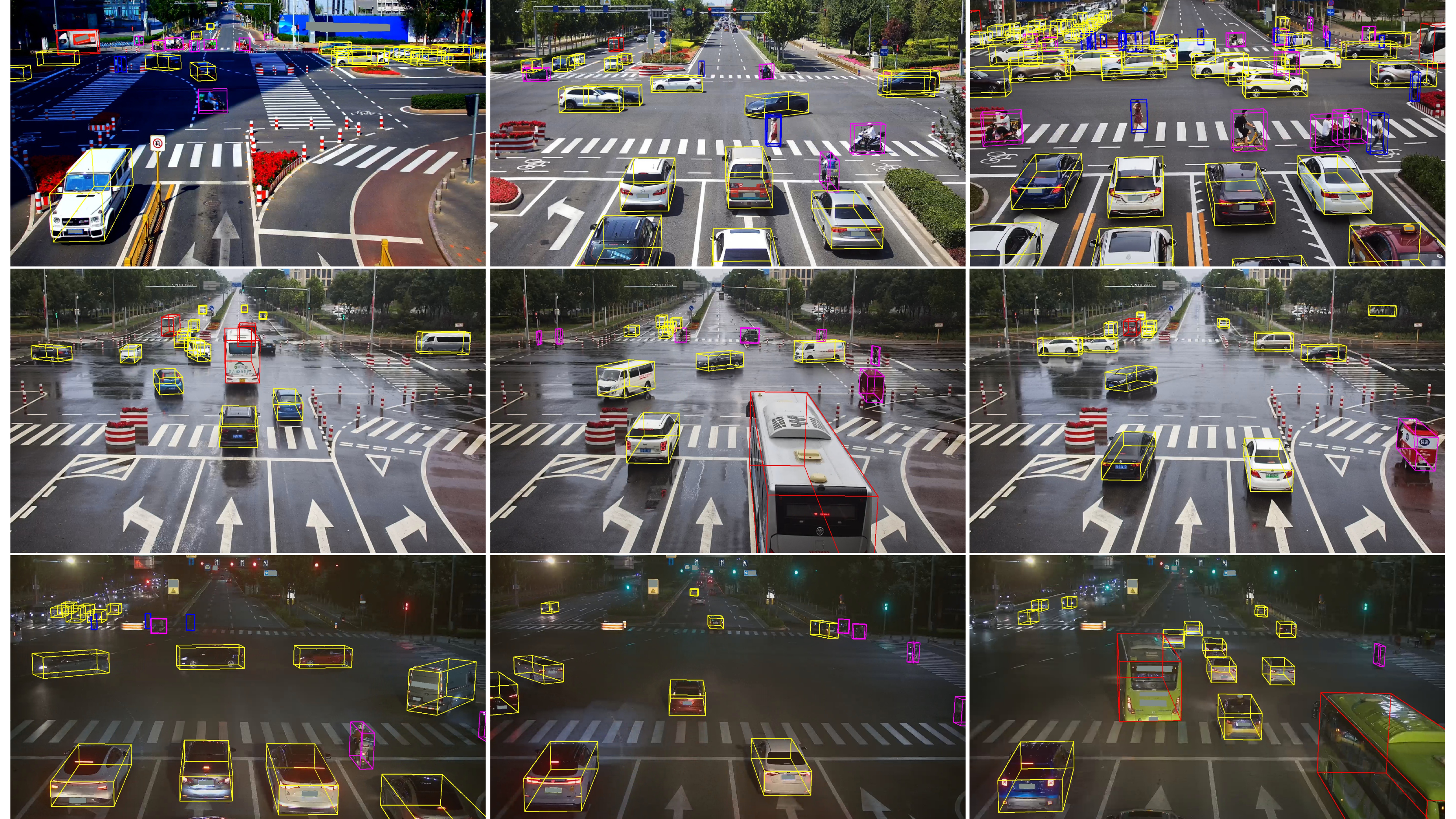}
 \end{center}
 \caption{More collected examples. From top to bottom, each row corresponds to clear/sunny/cloudy, rainy and dawn/dusk.}
 \label{Figure1_data_visualization}
\end{figure*}

We present more roadside data samples for visualization in Fig.~\ref{Figure1_data_visualization}, including different weather conditions, collecting time and object densities.

\begin{table*}[t]
\begin{center}
  \resizebox{\linewidth}{!}{  
\begin{tabular}{c| c | c | c | cc cc |cc cc }
\specialrule{0.8pt}{0.1pt}{0pt}
\multirow{3}{*}{Setting} & \multirow{3}{*}{Method} & \multirow{3}{*}{Backbone} &\multirow{3}{*}{Branch} & \multicolumn{4}{c|}{IoU = 0.25} & \multicolumn{4}{c}{IoU = 0.5} \\ 
&  &  &   & \multicolumn{2}{c}{Cyclist} & \multicolumn{2}{c|}{Pedestrian} & \multicolumn{2}{c}{Cyclist} & \multicolumn{2}{c}{Pedestrian} \\ 
&  &  &   & AP$_{\text{3D}{|\text{R40}}}$ & Rope$_\text{score}$ & AP$_{\text{3D}{|\text{R40}}}$ & Rope$_\text{score}$  & AP$_{\text{3D}{|\text{R40}}}$ & Rope$_\text{score}$ & AP$_{\text{3D}{|\text{R40}}}$ & Rope$_\text{score}$\\ 
\hline
\multirow{5}{*}{$\mathbb{I}$} 	
& M3D-RPN-$\bm{(G)}$~\cite{brazil2019m3d} & \multirow{1}[0]{*}{ResNet34}  & A 
&12.45 & \multicolumn{1}{c|}{28.64}	& 2.29 &  20.07 
&2.61 & \multicolumn{1}{c|}{20.79} & 0.34  &  18.63 \\

&  M3D-RPN-$\bm{(D)}$~\cite{brazil2019m3d} & \multirow{1}[0]{*}{ResNet34} 	& A 
&22.26 & \multicolumn{1}{c|}{36.61}	& 6.98 &  24.00 
&5.64  & \multicolumn{1}{c|}{23.35} & 1.16  &  19.47 \\

&  Kinematic3D-$\bm{(G)}$~\cite{brazil2020kinematic} & \multirow{1}[0]{*}{DenseNet121} & A	 
&14.78 & \multicolumn{1}{c|}{29.72}	& 3.59 &  21.19 
&2.97 & \multicolumn{1}{c|}{2.34} & 0.52  &  18.92 \\
 &  MonoDLE-$\bm{(G)}$~\cite{ma2021delving} & \multirow{1}[0]{*}{DLA-34} & K  
 & 24.26 & \multicolumn{1}{c|}{37.35}	& 4.14 &  21.85 
&4.68 & \multicolumn{1}{c|}{21.70} & 0.44  &  18.91 \\

 & MonoFlex-$\bm{(G)}$~\cite{zhang2021objects} & \multirow{1}[0]{*}{DLA-34}  & K	 
 & 65.63 & \multicolumn{1}{c|}{70.78}	& 36.83 &  48.10 
&24.25 & \multicolumn{1}{c|}{37.70} & 7.58  &  24.70 \\

\hline
\hline
\multirow{5}{*}{$\mathbb{II}$} 	
& M3D-RPN-$\bm{(G)}$~\cite{brazil2019m3d} & \multirow{1}[0]{*}{ResNet34}  & A  
& 5.07 & \multicolumn{1}{c|}{22.42}	& 1.40 &  19.40 
&0.75 & \multicolumn{1}{c|}{19.02} & 0.25  &  18.54 \\

&  M3D-RPN-$\bm{(D)}$~\cite{brazil2019m3d} & \multirow{1}[0]{*}{ResNet34} 	& A 
& 11.22 & \multicolumn{1}{c|}{27.54}	& 3.93 &  21.54 
&2.09 & \multicolumn{1}{c|}{20.25} & 0.67  &  19.08 \\

&  Kinematic3D-$\bm{(G)}$~\cite{brazil2020kinematic} & \multirow{1}[0]{*}{DenseNet121} & A	 
& 4.84 & \multicolumn{1}{c|}{21.15}	& 2.98 &  20.52 
&0.72 & \multicolumn{1}{c|}{17.94} & 0.73  &  19.02 \\
 &  MonoDLE-$\bm{(G)}$~\cite{ma2021delving} & \multirow{1}[0]{*}{DLA-34} & K  
 & 10.93 & \multicolumn{1}{c|}{26.44}	& 3.72 &  21.42 
&2.02 & \multicolumn{1}{c|}{19.32} & 0.47  &  18.86 \\

 & MonoFlex-$\bm{(G)}$~\cite{zhang2021objects} & \multirow{1}[0]{*}{DLA-34}  & K	
 & 44.27 & \multicolumn{1}{c|}{53.58}	& 25.48 &  39.04 
&12.30 & \multicolumn{1}{c|}{28.00} & 4.29  & 22.09  \\

\specialrule{0.7pt}{0.1pt}{0pt}
\end{tabular}
}
\end{center}
\vspace*{-3mm}
\caption{Performance of the Pedestrian and Cyclist on the Rope3D Dataset with IoU = 0.25 and 0.5 under two train-val splitting settings: the homologous ($\mathbb{I}$) and the heterologous ($\mathbb{II}$). -$\bm{(G)}$ denotes adapting the ground plane, -$\bm{(D)}$ means using the depth map of ground. The abbr. in the branch column denotes: A: anchor-based, K: keypoint-based.}
\label{tab_performance_overall_add}
\end{table*}
\begin{table*}[t]
\vspace*{-1mm}
\begin{center}
  \resizebox{\textwidth}{!}{  
\begin{tabular}{l | c | c | c c c  c c c c c c }
\specialrule{0.8pt}{0.1pt}{0pt}
\multirow{2}{*}{Setting} & \multirow{2}{*}{Method} & \multirow{2}{*}{Backbone} & \multicolumn{8}{c}{AP$_{\text{3D}{|\text{R40}}}$[Mod] / Rope$_\text{score}$ } &  \\ 
& & & car & van & bus & truck & cyclist & motorcyclist & tricyclist & pedestrian  \\ 
\hline
\multirow{5}{*}{$\mathbb{I}$} & 
\multirow{1}{*}{M3D-RPN-$\bm{(G)}$~\cite{brazil2019m3d} } 		&  ResNet34	& 41.15 / 52.38	& 31.19 / 44.31	& 32.60 / 44.58	& 26.54 / 39.89	& 6.48 / 23.80	& 10.23 / 26.84	& 20.81 / 35.39	& 2.01 / 19.87  	\\ 
& 
\multirow{1}{*}{M3D-RPN-$\bm{(D)}$~\cite{brazil2019m3d}} 		&  ResNet34	& 64.38 / 71.04	& 48.56 / 58.33	& 41.67 / 52.06	& 39.14 / 50.09	& 16.64 / 32.11	& 24.46 / 38.41	& 41.77 / 52.40	& 6.22 / 23.42  	\\ 
& 
\multirow{1}{*}{Kinematic3D-$\bm{(G)}$~\cite{brazil2020kinematic}}& DenseNet121 	& 48.42 / 57.32	& 34.13 / 45.86	& 21.71 / 35.43	& 32.30 / 43.46	& 8.45 / 24.84	& 18.66 / 32.77	& 28.66 / 40.99	& 3.25 / 20.90 	\\ 
& 
\multirow{1}{*}{MonoDLE-$\bm{(G)}$~\cite{ma2021delving}} 		&  DLA-34	& 77.76 / 81.11	& 67.52 / 72.74	& 66.24 / 71.57	& 54.74 / 61.33	& 58.64 / 65.27	& 65.51 / 70.55	& 73.62 / 77.12	& 41.68 / 52.02  	\\ 
& 
\multirow{1}{*}{MonoFlex-$\bm{(G)}$~\cite{zhang2021objects}} 	& DLA-34 	& 51.89 / 60.41	& 54.18 / 62.02	& 47.17 / 56.24	& 53.18 / 60.74	& 58.41 / 65.37	& 67.30 / 72.10	& 69.67 / 73.74	& 26.72 / 40.02  	\\ 

\hline
\hline
\multirow{5}{*}{$\mathbb{II}$} & 
\multirow{1}{*}{M3D-RPN-$\bm{(G)}$~\cite{brazil2019m3d} } 		&  ResNet34	 & 15.51 / 31.51	& 5.96 / 23.65	& 23.74 / 37.29	& 7.50 / 23.94	&  1.79 / 19.83	&  3.41 / 20.87	& 10.75 / 27.14	& 1.78 / 19.62   	\\ 
& 
\multirow{1}{*}{M3D-RPN-$\bm{(D)}$~\cite{brazil2019m3d}} 		&  ResNet34	& 34.25 / 46.74	& 22.45 / 37.18	& 57.90 / 65.04	& 27.30 / 40.42	& 21.58 / 36.00	& 15.08 / 30.58	& 19.75 / 34.80	& 5.14 / 22.53 \\ 
& 
\multirow{1}{*}{Kinematic3D-$\bm{(G)}$~\cite{brazil2020kinematic}}& DenseNet121 	&   22.38 / 36.20	& 10.13 / 26.42	& 22.25 / 35.34	& 9.86 / 25.33	& 2.54 / 19.79	& 5.52 / 21.56	& 14.25 / 29.98	& 1.66 / 19.33 	\\ 
& 
\multirow{1}{*}{MonoDLE-$\bm{(G)}$~\cite{ma2021delving}} 		&  DLA-34	&  25.78 / 39.30	& 15.80 / 31.00	& 60.22 / 66.26	& 16.47 / 30.20	& 25.25 / 38.38	& 23.86 / 37.07	& 26.80 / 39.96	& 30.70 / 43.14 \\ 
& 
\multirow{1}{*}{MonoFlex-$\bm{(G)}$~\cite{zhang2021objects}} 	& DLA-34 	& 24.44 / 38.41	& 16.36 / 31.51	& 41.09 / 50.21	& 26.35 / 39.20	& 47.26 / 56.25	& 51.55 / 59.22	& 18.32 / 33.32	& 22.96 / 37.11  \\ 

\specialrule{0.8pt}{0.1pt}{0pt}
\end{tabular}
}
\end{center}
\vspace*{-3mm}
\caption{Performance of the fine-grained categories on the Rope3D Dataset under the homologous ($\mathbb{I}$) and heterologous ($\mathbb{II}$) settings, respectively. -$\bm{(G)}$ denotes adapting the ground plane, -$\bm{(D)}$ means using the depth map of ground plane. The abbr. in the branch column denotes: A: anchor-based, K: keypoint-based. IoU = 0.25 for non-motor vehicles and pedestrian, IoU = 0.5 for motor vehicles.}
\label{9categories}
\end{table*}

\begin{table*}[t]
\begin{center}
  \resizebox{\linewidth}{!}{  
\begin{tabular}{c| c | cc cc |cc cc }
\specialrule{0.8pt}{0.1pt}{0pt}
\multirow{3}{*}{Setting} & \multirow{3}{*}{Method}  & \multicolumn{8}{c}{AP$_{\text{3D}{|\text{R40}}}$[Mod] / Rope$_\text{score}$ } \\ 
&    & \multicolumn{2}{c}{IoU = 0.5} & \multicolumn{2}{c|}{IoU = 0.7} & \multicolumn{2}{c}{IoU = 0.25} & \multicolumn{2}{c}{IoU = 0.5} \\ 
&    & Car & Big Vehicle & Car & Big Vehicle  & Cyclist & Pedestrian & Cyclist & Pedestrian \\ 
\hline
\multirow{4}{*}{$\mathbb{I}$} 	
& MonoDLE-$\bm{(G)}$~\cite{ma2021delving} &  
 51.70 / 60.36 & \multicolumn{1}{c|}{40.34 / 50.07}&	
 13.58 / 29.46 & 9.63 / 25.80  
 & 24.26 / 37.35 &\multicolumn{1}{c|}{4.14 / 21.85} &   
 4.68 / 21.70 & 0.44 / 18.91 \\
 
 & MonoDLE-$\bm{(D)}$~\cite{ma2021delving}  & 
  77.50 / 80.84 & \multicolumn{1}{c|}{49.07 / 57.22}&	
 54.53 / 62.48 & 17.25 / 32.00  
 & 61.81 / 67.57 &\multicolumn{1}{c|}{35.72 / 47.22} &   
 32.60 / 44.22 & 12.96 / 29.03 \\
 
 & MonoFlex-$\bm{(G)}$~\cite{zhang2021objects} &  
 60.33 / 67.86 & \multicolumn{1}{c|}{37.33 / 47.96}&	
 33.78 / 46.12 & 10.08 / 26.16  
 & 65.63 / 70.78 &\multicolumn{1}{c|}{36.83 / 48.10} &   
 24.25 / 37.70 & 7.58 / 24.70 \\
 
 & MonoFlex-$\bm{(D)}$~\cite{zhang2021objects}  & 
  59.78 / 66.66 & \multicolumn{1}{c|}{59.81 / 66.07}&	
  35.64 / 47.43 & 24.61 / 38.01 
 & 74.09 / 77.45 &\multicolumn{1}{c|}{50.46 / 59.03} &   
 39.33 / 49.64 & 13.55 / 29.50 \\

\hline
\hline
\multirow{4}{*}{$\mathbb{II}$} 	
& MonoDLE-$\bm{(G)}$~\cite{ma2021delving} &  
 19.08 / 33.72 & \multicolumn{1}{c|}{19.76 / 33.07}&	
 3.77 / 21.42 & 2.31 / 19.55  
 & 10.93 / 26.44 &\multicolumn{1}{c|}{3.72 / 21.42} &   
 2.02 / 19.32 & 0.47 /18.86 \\
 
 & MonoDLE-$\bm{(D)}$~\cite{ma2021delving}  & 
 31.33 / 43.68 & \multicolumn{1}{c|}{23.81 / 36.21}&	
 12.16 / 28.39 & 3.02 / 19.96
 & 27.59 / 39.83 &\multicolumn{1}{c|}{25.33 / 38.82} &   
 10.00 / 25.78 & 7.31 / 24.45 \\
 
 & MonoFlex-$\bm{(G)}$~\cite{zhang2021objects} & 
  32.01 / 44.37 & \multicolumn{1}{c|}{13.86 / 28.47}&	
 10.86 / 27.39 & 0.97 / 18.18  
 & 44.27 / 53.58 &\multicolumn{1}{c|}{25.48 / 39.04} &   
 12.30 / 28.00 & 4.29 / 22.10 \\
  & MonoFlex-$\bm{(D)}$~\cite{zhang2021objects} &  
   37.27 / 48.58 & \multicolumn{1}{c|}{47.52 / 55.86}&	
  11.24 / 27.79 & 13.10 / 28.22  
 & 40.78 / 50.62 &\multicolumn{1}{c|}{ 37.79 / 48.91} &   
 13.64 / 28.93 & 7.53 / 24.72 \\

\specialrule{0.7pt}{0.1pt}{0pt}
\end{tabular}
}
\end{center}
\vspace*{-3mm}
\caption{Performance of the vanilla and improved approaches on the Rope3D Dataset under two train-val splitting settings: the homologous ($\mathbb{I}$) and the heterologous ($\mathbb{II}$). -$\bm{(G)}$ denotes adapting the ground plane, -$\bm{(D)}$ means using the depth map of ground.}
\label{tab_performance_monoflex}
\end{table*}

\section{Additional Experiments}
\label{sec:add_experiments}
In this section, we show more experimental results. As is stated, we offer two kinds of validation sets, the homologous ($\mathbb{I}$) in which the training and validation set have common scenes, and the heterologous ($\mathbb{II}$) with the validation set has never seen the scenes in the training set and the camera specifications are possibly different with the training set.

\noindent{\textbf{Performance of pedestrian and cyclist.}}
In addition to the results of motor vehicles in the main paper, we further present the results of pedestrians and cyclists under the homologous and heterologous settings in Table~\ref{tab_performance_overall_add} for further evaluation. Monoflex~\cite{zhang2021objects} obtains superior performance especially on cyclist and pedestrian categories, which shows consistent behavior with the original work.

\noindent{\textbf{Performance of fine-grained categories.}}
We conduct two levels of categorization and the corresponding experiments. For the coarse-grained level, the monocular 3D object detection task mainly focuses on the most common traffic elements: Car, Big Vehicle, Pedestrian, and Cyclist. For fine-grained level, Car includes car and van, Big Vehicle can be further divided into truck and bus, and meanwhile, Cyclist can be subdivided into cyclist, motorcyclist, and tricyclist as they are driving non-motor vehicles.
The performances of fine-grained-level-8 are compared in Table~\ref{9categories}.

 \noindent{\textbf{Performance of vanilla and improved approaches.}} 
 Leveraging the depth map of ground plane, we try to alleviate the ambiguity caused by different camera specifications. For this purpose, we evaluate two approaches to incorporate depth information with the RGB appearance feature. The first one is directly concatenating the depth map with the original RGB channels as input, and the second is adopting another siamese network for depth feature extraction and further weighted fusion of the two depth predictions. The performances of these two methods are similar and we hence only report the results by concatenation, a simple yet effective improvement strategy. We believe more sophisticated approaches might further improve the performance, which is out of the scope of this paper.
In addition to the reported results of M3D-RPN (anchor-based) in the main paper, we also apply the depth map of the ground plane to MonoDLE and MonoFlex, the keypoint-based approaches. The comparison results are presented in Table~\ref{tab_performance_monoflex}. Comparing the vanilla and improved approaches, a consistent performance gain has been observed across all the baselines.

{\small
\bibliographystyle{ieee_fullname}
\bibliography{newbib}
}
\newpage

\end{document}